\documentclass[manuscript,screen,nonacm]{acmart}  
  
\usepackage{amsmath}  
\usepackage{multirow}  
\usepackage{enumitem}  
\usepackage{subcaption}  
\usepackage{float} 
\usepackage{placeins}  
\usepackage{tikz}  
  
\usepackage[capitalize]{cleveref}  
  
\hypersetup{  
  colorlinks=true,  
  linkcolor=blue!60!black,  
  citecolor=blue!60!black,  
  urlcolor=blue!60!black  
}

\title{Referential Uncertainty in Human--AI Collaboration}  
  
\author{Christian Poelitz}
\affiliation{%
  \institution{Microsoft Research}
  \city{Cambridge}
  \country{UK}
}
\email{christian.poelitz@microsoft.com}

\author{Finale Doshi-Velez}
\affiliation{%
  \institution{Harvard University}
  \city{Cambridge}
  \state{MA}
  \country{USA}
}
\email{finale@seas.harvard.edu}

\author{Si\^{a}n Lindley}
\affiliation{%
  \institution{Microsoft Research}
  \city{Cambridge}
  \country{UK}
}
\email{sian.lindley@microsoft.com}

\begin{document}  
\begin{abstract}
Effective human--AI collaboration often requires partners to establish references through interaction. This referential grounding becomes fragile when descriptions are ambiguous, similar or distracting referents compete, or partners see different things. We study \emph{referential uncertainty}---uncertainty over which candidate object a description refers to---in a collaborative puzzle task in which a human Helper instructs an AI Worker to place pieces. In this context, the Worker needs to \emph{identify} and \emph{communicate} its uncertainty, while the human Helper needs to \emph{recognize} and \emph{act} on it. In a real human--AI puzzle task, we show that a separately elicited belief distribution over candidate puzzle pieces is better calibrated (ECE $0.15$) and better discriminates correct from incorrect placements (AUROC $0.65$) than raw action-token probabilities, which are severely overconfident ($0.97$ mean confidence and ECE $0.44$). Across three frontier vision--language models (\textsc{GPT-4.1}, \textsc{GPT-5}, \textsc{GPT-5.5}), we show empirically that this elicited uncertainty increases predictably with instruction vagueness, but not with alternative referents in context, even when those alternatives increase errors. The models seldom externalize this uncertainty, asking for clarification on only $3.5$--$16.7\%$ of instruction turns. In a controlled human study ($N=210$), we show that humans given only the AI Worker's default message accept $78\%$ of wrong placements and cannot distinguish right from wrong (AUC $0.50$). Precise descriptions and, especially, well-targeted hedges reduce acceptance of wrong moves to $36\%$, while largely preserving acceptance of correct moves, thereby compensating for missing shared awareness, such as not seeing the Worker's action directly. However, this benefit depends on targeting quality: a realistic hedge derived from the model's own belief entropy inherits weaknesses in that signal and can do more harm than good. Externalized uncertainty therefore helps a human partner only when it is accurately targeted.\end{abstract}

\received{20 February 2007}
\received[revised]{12 March 2009}
\received[accepted]{5 June 2009}

\maketitle

\section{Introduction}  
\label{sec:intro}  
Collaborative tasks often require partners to identify and manipulate relevant objects through verbal descriptions or nonverbal cues, such as pointing. For example, solving a puzzle together requires partners to establish which pieces each person is referring to. This can be challenging when descriptions are ambiguous (e.g., ``the pink piece''), multiple objects match a description, or partners have asymmetric information (e.g., each partner sees only part of the puzzle). Work by Clark, Brennan, and colleagues~\cite{clark1986referring, brennan1996conceptual} examines how collaborative partners establish mutual understanding through \emph{grounding}: an interactive process of building common ground. In referential communication, one partner presents a referring expression, and the partners may use clarification questions, corrections, and other evidence of understanding to establish the intended referent. Acceptance provides evidence that their understanding is sufficient for the current purpose, although it may need to be revised as the interaction proceeds.

Human--AI collaboration presents similar challenges. A human and an AI agent must establish common ground about their goals, instructions, and the objects or actions under discussion. In this work, we focus on human--AI collaboration as an instance of \emph{ad hoc teamwork}~\cite{stone2010ad, altschuller2010trust}, 
in which partners have not trained together and begin with limited shared experience. Unlike established teams, which can develop shared conventions and familiarity with one another's capabilities, ad hoc partners must establish this understanding during the interaction. In human--AI collaboration, limited prior experience may also make it difficult for the human partner to judge the AI's capabilities and recognize when its judgments require scrutiny~\cite{li2026learning}.  
  
The present work investigates \emph{referential uncertainty} in a human--AI collaborative puzzle task~\cite{poelitz2026benchmarkassesscommonground}. A human Helper sees a target puzzle configuration, while an AI Worker sees a set of available pieces. The Helper instructs the Worker which pieces to place and where to place them. The Worker must identify and place the pieces intended by the Helper. Because the partners have asymmetric information, a description that seems sufficiently specific to the Helper may fail to distinguish the intended piece from plausible alternatives available to the Worker. For example, the Helper might instruct: ``Place the pink spiral piece at the top left.'' If the Worker sees two pieces matching this description, the instruction leaves two plausible referents. Resolving this referential uncertainty may require further grounding before the Worker can confidently proceed.  
  
We examine four capabilities involved in resolving such uncertainty: the Worker's ability to (a) identify referential uncertainty and (b) communicate it, and the Helper's ability to (c) recognize the uncertainty and (d) act on it to resolve the ambiguity. Existing research addresses related questions through uncertainty calibration, clarification-question generation, and human reliance on AI. Our focus, however, is on how these capabilities connect within collaborative referential communication. Specifically, we investigate uncertainty over alternative referents in the immediate task context; how and when an AI should communicate that uncertainty through clarification rather than simply abstaining; whether a human collaborator recognizes the uncertainty; and whether the collaborator can provide information that resolves it.  

\subsection{Contributions}  
We summarize our contributions below:  
\begin{itemize}  
    \item We study \emph{referential uncertainty} and grounding in human--AI collaboration, distinguishing four key capabilities: the Worker's skills in identifying and communicating this uncertainty, and the Helper's abilities in recognizing and resolving it. Building on the collaborative puzzle benchmark of~\cite{poelitz2026benchmarkassesscommonground}, we operationalize referential uncertainty as a distribution over candidate referents and design a framework to measure it using three signals---raw and temperature-calibrated action-token log-probabilities and a separately elicited (verbalized) belief distribution.  
      
    \item Through extensive experiments with three state-of-the-art vision--language models (\textsc{GPT-4.1}, \textsc{GPT-5}, \textsc{GPT-5.5}), we assess AI agents' abilities to represent and express referential uncertainty. Our findings highlight that a verbalized belief distribution is better calibrated than raw probabilities, particularly as instruction vagueness increases, while context modality (visual vs.\ textual) has only small, model-dependent effects and confusable alternative referents barely move it even as they drive error. Despite these findings, models rarely externalize this uncertainty, seldom asking for clarification and instead echoing the Helper's brevity.  
      
    \item An exploratory user study ($N\!=\!210$) examines human perceptions of AI Worker confidence and accuracy in a collaborative puzzle task. The study reveals that shared awareness---seeing the Worker's board or a textual description of its chosen piece---and explicit uncertainty signals (hedging) influence perceived confidence and decision-making, though these signals must be accurately targeted to be beneficial. Inspecting whether the model's own elicited uncertainty can serve as that externalization signal, we find that this imperfect signal can do more harm than good in human--AI collaboration. 
\end{itemize}  

\subsection{Results}
On a collaborative puzzle task, we find that AI agents can represent referential uncertainty but systematically fail to express it. We empirically show that among three uncertainty signals, raw action-token log-probabilities are severely over-confident (mean confidence $0.97$ at $53\%$ accuracy; ECE $=0.44$), whereas a separately elicited (verbalized) belief distribution over referents is well calibrated and the best discriminator of correct from incorrect placements (AUROC $=0.65$, ECE $=0.15$). Across three frontier vision--language models (\textsc{GPT-4.1}, \textsc{GPT-5}, \textsc{GPT-5.5}), we show that the elicited uncertainty increases predictably with independent sources of ambiguity: vaguer instructions (belief entropy increases from $1.29$ to $2.77$ nats; error increases from $0\%$ to $77\%$ from over-specified to vague), while confusable alternatives in context increase error (from $19\%$ to $50\%$) but barely move the elicited uncertainty. The modality---textual or visual context---has less effect, especially for more recent models.  
  
The models rarely act on their uncertainty---performing actions on most of their most-uncertain turns resulting in low accuracy (belief-entropy AUROC $0.55$--$0.66$), asking for clarification on only $3.5\%$ (\textsc{GPT-4.1}) to $16.7\%$  (\textsc{GPT-5}) of instruction turns (image modality), and echoing the Helper's brevity instead of describing their choice. A controlled human study ($N=210$) confirms that these signals are what collaborative partners need: with only default messaging and limited awareness of the AI's context, humans accept $78\%$ of wrong placements and cannot tell right from wrong (AUC $=0.50$). However, when the AI describes its action in detail and externalizes its uncertainty, wrong-move acceptance falls to $36\%$ (with correct-move acceptance largely preserved, $74\%$), and partners with limited awareness detect errors about as well as those who can see the AI's actions.  

This benefit depends on \emph{which} turns are hedged: the AI can hedge only from its own belief entropy, which flags errors weakly (AUROC  
$\approx 0.55$). Following such poorly targeted hedges, partners defer to the uncertainty without making better decisions---the benefit comes from accurate targeting, not from the hedge itself.

\section{Prior Work}
\label{sec:prior_work}  
Our work draws on research in several key areas: common ground and grounding, trust and appropriate reliance, and uncertainty estimation and calibration. Below, we review how conversational partners establish shared understanding, how uncertainty expressions influence reliance on AI, and how AI uncertainty can be estimated and communicated.   
  
\subsection{Common Ground and Grounding}  
Building common ground is essential to collective action and increasingly recognized as important to human--AI interaction. \citet{Clark1991GroundingIC} describe grounding as a collaborative process through which participants establish sufficient mutual understanding for their conversational purposes. This involves coordinating both content---shared knowledge, beliefs, and assumptions---and process, including identifying and repairing misunderstandings. Participants present and accept information, continually updating common ground through cues such as acknowledgments, clarification requests, relevant responses, and continued attention.  
  
Grounding follows the principle of least collaborative effort: participants seek to minimize the joint work required to achieve mutual understanding. Rather than producing a perfect message in a single utterance, speakers may offer provisional descriptions or divide complex information into installments, checking understanding along the way. Grounding therefore requires both speakers and listeners to signal, monitor, and repair understanding.  
  
Research demonstrates how grounding improves communication efficiency. Krauss and colleagues  \citep{krauss1966concurrent,krauss1969development} found that participants describing novel patterns developed shared references, allowing them to use fewer words over time. \citet{CLARK19861} emphasize that these references are collaboratively negotiated through clarification and repair before becoming accepted common ground. As shared referential language develops, participants require fewer words and fewer clarification or repair exchanges. \citet{fussell1992coordination} also show that participants often begin with minimal descriptions and partially inaccurate assumptions about one another's knowledge, using feedback to achieve sufficient understanding. Once grounded, however, shared references tend to remain stable, even when more efficient alternatives are available \citep{brennan1996conceptual}.  
  
In human--AI collaboration, communicating uncertainty can help participants identify gaps in understanding and determine when clarification is needed. Its effectiveness depends both on the AI's ability to express uncertainty reliably and on the human's ability to interpret and act on those signals. Recent work has extended this research to grounding with LLMs~\cite{shaikh-etal-2024-grounding, shaikh2025navigatingriftshumanllmgrounding} and their use of clarification questions~\cite{poelitz2026teacher}, examining these issues in collaborative settings~\cite{collabllm2025} and cases of referential ambiguity~\cite{madge2025referential}.
  
\subsection{Trust and Appropriate Reliance}  
  
Related research has examined uncertainty communication primarily through its effects on human trust and reliance, rather than through the collaborative process of grounding. In particular, prior work has investigated how confidence expressions and explanations influence whether users accept AI recommendations.  
  
\citet{zhang2020effect} found that displaying task-specific confidence scores made users more likely to accept high-confidence recommendations than low-confidence ones. However, this did not improve decision quality because human and AI capabilities were not complementary: users also struggled with the cases in which the AI had low confidence. Similarly, \citet{bansal2021does} found that explanations could encourage users to accept AI recommendations without improving performance beyond that achieved by displaying confidence estimates alone, limiting the benefits of human--AI collaboration.  
  
\citet{rechkemmer2022confidence} showed that stated confidence influences users' beliefs about the correctness of AI responses, but observed accuracy has a stronger effect on acceptance. This suggests that feedback about AI performance can be more influential than stated confidence, even when that confidence is accurate. \citet{schemmer2023appropriate} examine \emph{appropriate reliance}: accepting correct AI recommendations and rejecting incorrect ones. They found that explanations increased confidence in AI and improved acceptance of correct recommendations, but did not improve rejection of incorrect recommendations.  
  
The form of uncertainty expression also matters. \citet{kim2024m} compared confident responses with hedged responses such as ``I'm not sure \ldots''. Hedging reduced users' confidence and acceptance rates without improving accuracy, suggesting that uncertainty expressions can shift reliance broadly without helping users distinguish correct from incorrect responses. Together, these findings highlight that uncertainty communication should support selective, appropriate reliance rather than simply increase or decrease trust.  
  
\subsection{Uncertainty Estimation and Calibration}  
  
Reliable uncertainty communication requires meaningful estimates of model uncertainty. Prior work has studied how to obtain these estimates and how they relate to AI decision-making. See \citet{xia2025survey} for a broader overview. Common approaches use token-level probabilities, variation across sampled responses, semantic entropy, or elicited verbal confidence.  
  
Token-level probabilities provide a relatively inexpensive basis for uncertainty estimation when they are available. \citet{kadavath2022language} showed that these probabilities can be well calibrated on some multiple-choice tasks. However, calibration is not consistent across settings \citep{lovering2025language}, and token-level probabilities may be less reliable for recent reasoning models \citep{guo2025deepseek}. Explicit calibration methods can address these limitations, although their applicability depends on the availability of suitable calibration data \citep{guo2017calibration}. Moreover, well-calibrated estimates must still be communicated effectively \citep{steyvers2025miscalibrated}, and their expression may depend on model-internal thresholding strategies \citep{kumaran2026causal}.  
  
Sampling-based approaches estimate uncertainty from variation across multiple generated responses without requiring access to token-level probabilities. \citet{kuhn2023semantic, Farquhar2024} combine sampling with semantic clustering to distinguish variation in meaning from variation in wording. These approaches offer an alternative basis for uncertainty estimation but incur additional computational cost.  
  
A further approach elicits confidence directly by asking models to report how likely their answers are to be correct. \citet{lin2022tmlr-teaching} showed that models can be fine-tuned to express confidence estimates, while subsequent work has explored eliciting confidence from existing models through prompting. Recent evidence suggests that models are becoming more capable of verbalizing and internally representing their confidence \citep{kumaran2026causal}. \citet{tian2023just} found that, for models fine-tuned with reinforcement learning from human feedback (RLHF), verbalized confidence can be better calibrated than raw token probabilities, particularly on multiple-choice tasks.  
  
However, elicited confidence should be distinguished from the certainty conveyed in an ordinary answer. \citet{zhou2024relying} argue that RLHF can encourage overly confident answers by penalizing expressions of uncertainty, even when confidence is not explicitly requested. Finally, \citet{xiong2024can} showed that elicited confidence estimates can be improved by prompting models to reason or by generating multiple answers and aggregating their agreement. These findings underscore that estimating uncertainty and expressing it appropriately are related but distinct challenges for human--AI collaboration.

\section{Task and Dataset}  
\label{sec:task}  

We study a collaborative puzzle~\cite{poelitz2026benchmarkassesscommonground} task in which a Helper guides a Worker to recreate a four-block target pattern using 24 visually ambiguous pieces. Only the Helper sees the target, while only the Worker sees the full piece set, requiring them to establish shared descriptions and coordinate placement. Across four trials (after a practice trial), the same target pieces appear in different configurations, allowing shared referring conventions and efficiency gains to develop. The benchmark varies whether the Helper can see the Worker’s workspace and assigns the AI to both roles to examine how visual context and role affect conversational grounding and task performance (see \cref{tab:dataset_stats}). In this work, we focus exclusively on tasks in which the Helper \textbf{cannot see the Worker’s actions}\footnote{We make the Worker’s actions visible only in a controlled condition in the human study of uncertainty and correctness perception described later.}, as these tasks showed frictions in human–AI collaboration as reported by \citet{poelitz2026benchmarkassesscommonground}: the AI Worker rarely asked clarification questions or expressed uncertainty, while the human Helper made very little correction or clarification attempts either.

Our motivating observation was that AI Workers often described their selected pieces vaguely, sometimes simply repeating the Helper's ambiguous description. Such responses provided little evidence that the intended piece had been selected, yet expressed no uncertainty that might prompt the Helper to check or intervene. This raises a central question: what could the Worker communicate differently to help its partner recognize and repair misunderstandings? We investigate two possibilities: making piece descriptions \textbf{more informative} and adding \textbf{explicit uncertainty signals}. Based on these observations, we test how descriptions and uncertainty cues affect human judgments of correctness across five conditions.

\subsection{Dataset}
We use the recorded interactions from the collaborative puzzle benchmark of \citet{poelitz2026benchmarkassesscommonground}. Our analyses focus  
on turns in which a human Helper instructs an AI Worker to select and place a puzzle piece. Each turn provides the conversation history including verbalized confirmations of actions like 'I placed the pieces to the top left', the  current Helper instruction, the Worker's available pieces and workspace, and the recorded Worker response. We extend these records with manual annotations of the intended referent (See \cref{sec:ground_truth}).  
  
We use these data in two complementary ways. First, we conduct offline model evaluations to compare uncertainty signals, examine their sensitivity to referential difficulty, and assess whether models communicate uncertainty through clarification requests. These evaluations include both the original recorded interactions and single-prompt replays of individual turns (\cref{sec:models}). Second, we construct stimuli for a human  
perception study by varying the information available to observers while retaining the underlying recorded placements\footnote{Message length is matched within each message design and differs only by the description itself: the two original-message conditions are equal in length (\emph{Generic} $31.5$ vs.\ \emph{Visible board} $30.4$ words; Holm $p{=}.53$) and the three description-based conditions lie within ${\sim}2$ words of one another (\emph{Described} $36.4$, \emph{Hedged} $37.7$, \emph{Self-hedged} $38.2$), so the only systematic gap is the ${\approx}5$ added description words separating the two groups (\emph{Described}$-$\emph{Generic} $+4.9$, $p{<}.001$)---the board-visibility and hedge effects are therefore not confounded with message length.}.
  
For the human study, we select interactions from the benchmark's non-shared-view condition, in which the Helper could not see the Worker's workspace. We restrict the rated turns to the first two puzzle trials. This subset captures interactions before partners have accumulated extensive experience with the recurring target pieces. Using the same underlying turns across conditions allows us to compare access to the Worker's board with changes to its verbal descriptions and uncertainty cues.

\section{Study design}
\label{sec:human_design}

\begin{figure}[t]
\centering
\includegraphics[width=\textwidth]{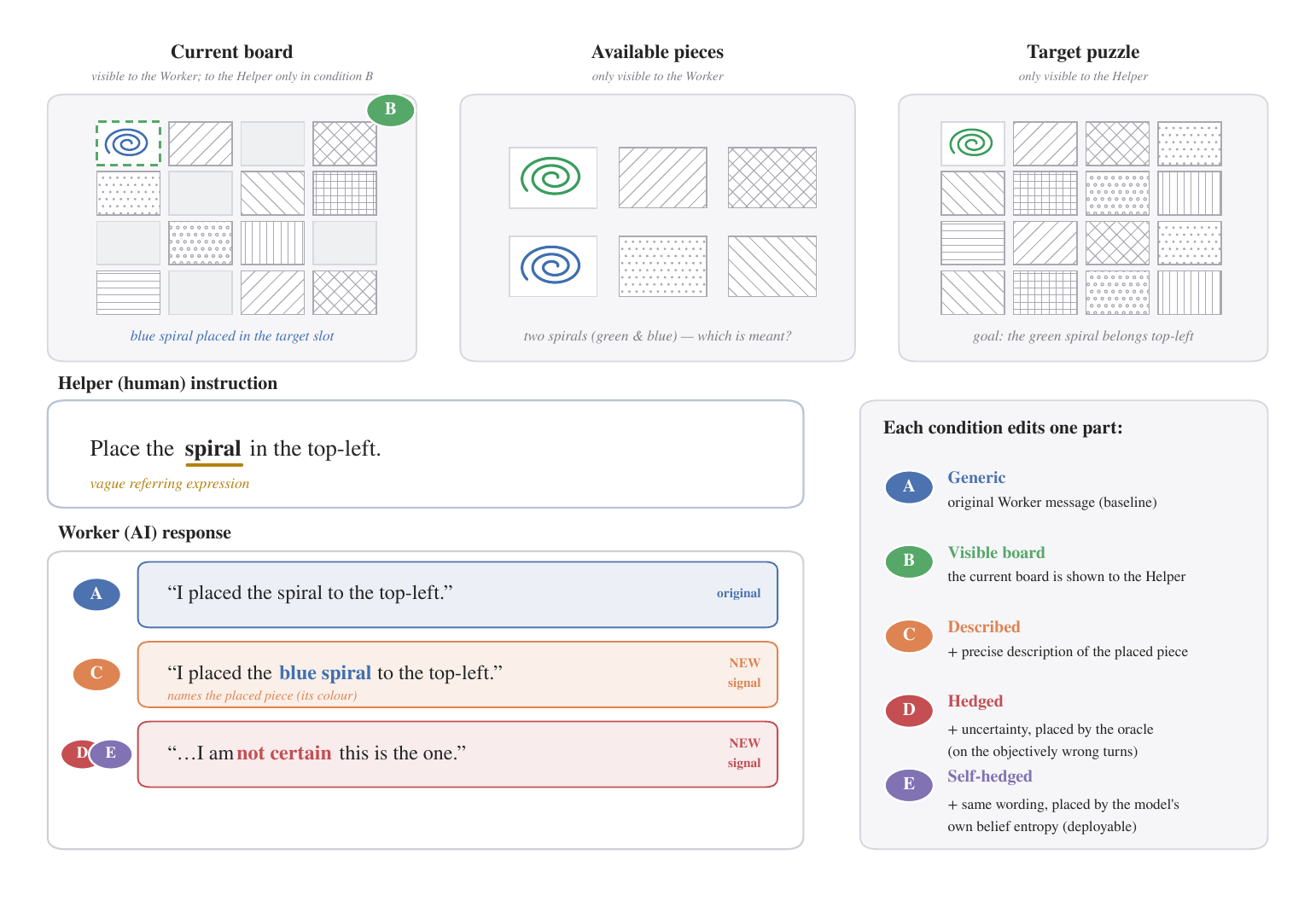}
\caption{\textbf{The collaborative puzzle task and the human study conditions.} The partners hold
asymmetric information: the \emph{Worker} (AI) sees the current board and all $24$ available
pieces but not the target, whereas the \emph{Helper} (human) sees the target pattern but
neither the pieces nor---by default---the Worker's board. The Helper gives a referring
instruction and the Worker selects, places a piece, and replies with a message (here, a blue
spiral placed in the target slot). Each condition changes exactly \emph{one} element of this
same turn. Four vary only the Worker's message while the board stays hidden:
\textbf{Generic}~(A) is the original message; \textbf{Described}~(C) appends a precise
description of the placed piece; and \textbf{Hedged}~(D) and \textbf{Self-hedged}~(E)
additionally append an expression of uncertainty---inserted by an oracle on the objectively
wrong turns (D) or, in the deployable variant, from the Worker's own belief entropy (E). The
fifth condition, \textbf{Visible board}, keeps the \emph{Generic} message but reveals the
Worker's board to the Helper---the shared-view control (\emph{A/shared}). Holding the turn
fixed and varying only the message or the board's visibility isolates how each externalized
signal shapes the Helper's perceived confidence, error detection, and willingness to act.}
\label{fig:task_schematic}
\end{figure}

We conduct a between-subjects study to examine how Worker messages and visual access affect human judgments of referential correctness. Participants viewed a recorded Helper--Worker interaction up to a selected Worker response and evaluated that response from the Helper's perspective. They rated the likelihood that the Worker had selected the intended piece, judged its apparent confidence, and indicated how they would respond.  

The study measures judgments and intended interventions rather than behaviour in a live collaboration: participants' choices did  
not change the recorded interaction or elicit a subsequent Worker response. (See Appendix for grading interface~\ref{fig:grader_ui})

\subsection{Conditions}  
\label{sec:human_conditions} 

\begin{table}[t]  
    \centering  
    \small  
    \caption{Human-study conditions. All conditions retain the  
    underlying recorded placement. The first four form the primary  
    comparison family; Self-hedged forms the secondary exploratory  
    family.}  
    \label{tab:conditions_primary}  
    \begin{tabular}{@{}lp{1.2cm}p{7.6cm}@{}}  
        \toprule  
        \textbf{Condition} & \textbf{Board} & \textbf{Worker message} \\  
        \midrule  
        Generic & Hidden &  
        Original Worker message. \\  
        Described & Hidden &  
        Original message plus a precise description of the  
        selected piece. \\  
        Hedged & Hidden &  
        Described message plus an uncertainty cue on  
        oracle-selected incorrect turns. \\  
        Visible board & Shown &  
        Original Worker message, with access to the Worker's  
        workspace. \\  
        \midrule  
        Self-hedged & Hidden &  
        Described message plus the same uncertainty cue on  
        turns selected using elicited belief entropy. \\  
        \bottomrule  
    \end{tabular}  
\end{table}
  
We compare five conditions that vary the information available about the Worker's selection (see \cref{tab:conditions_primary} for a summary). The underlying placement is held fixed. The manipulations change the Worker's message or the access to its workspace. See \cref{fig:task_schematic} for an illustration. 
  
\paragraph{Primary comparison family.}  
The \emph{Generic} condition presents the original Worker messages without revealing the Worker's board. The \emph{Described} condition adds a precise description of the piece actually selected, allowing to compare that selection with the Helper's instruction. The \emph{Hedged} condition adds explicit uncertainty cues to the detailed descriptions. These cues are oracle-targeted i.e., ground-truth correctness is used to place hedges on a subset of incorrect selections, covering approximately half of the incorrect turns. Finally, the \emph{Visible board} condition retains the original messages but reveals the Worker's workspace.  
  
Together, these conditions distinguish the effects of informative descriptions, targeted uncertainty cues, and direct visual access. Oracle-targeted hedging is an experimental intervention rather than a deployable policy, because it requires knowledge of whether a selection is incorrect.  

\paragraph{Secondary exploratory comparison family.}  
The \emph{Self-hedged} condition uses the same hedge wording as \emph{Hedged}, but selects turns using the Worker's separately elicited belief entropy rather than ground-truth correctness. It therefore tests a model-derived alternative that could be applied without access to the intended referent. Concretely, the hedge is placed whenever the Worker's elicited belief entropy exceeds a fixed threshold.

\subsection{Dependent Variables}  
\label{sec:human_measures}  
  
For each rated Worker turn, participants report:  
\begin{itemize}[leftmargin=*,nosep]  
    \item \textbf{Correctness likelihood} ($0$--$100$):  
    how likely it is that the Worker selected the intended piece.  
    \item \textbf{Apparent confidence} ($0$--$100$):  
    how confident the Worker sounds.  
    \item \textbf{Intended action}: proceed/accept, change/undo,  
    ask the Worker to explain or verify, or provide a  
    more-specific instruction.  
\end{itemize}  
  
Correctness likelihood and apparent confidence measure distinct judgments. A participant may perceive a response as confident without believing that it is correct. See Appendix \cref{fig:grader_ui} for the grading UI.

We summarize responses at the participant level and report additionally \emph{rating discrimination} and \emph{action discrimination} as described in \cref{sec:discriminations} .   
  
Additional measures include participant-level mean ratings, the distribution of intended actions, and questionnaire responses about prior attitudes toward AI and the perceived quality of  the interaction. We assess discrimination using participant-level AUC and calibration using correctness ratings rescaled to $[0,1]$.

\subsection{IRB}
The study protocol was reviewed and approved by the relevant Institutional Review Board (IRB). All participants provided informed consent and received a post-study debrief explaining the experimental conditions, including any modifications made to the AI-generated dialogue.

\subsection{Participants}
\label{sec:human_participants}

We recruited participants through Prolific (\url{https://www.prolific.com}). Eligible participants were at least $18$ years old, resident in the United Kingdom, and fluent in English, with no self-reported language-related, visual, or cognitive impairments. After excluding returned, timed-out, and rejected submissions ($37$ returned, $1$ timed out, $2$ rejected), $210$ participants completed the study (based on a power analysis) and were retained for analysis ($107$ male, $101$ female, $2$ undisclosed). Their ages ranged from $18$ to $76$ years (mean $37.5$, median $35$) Each participant was randomly assigned to a single condition, yielding $41$--$43$ participants per condition (Generic~$41$, Described~$42$, Hedged~$42$, Visible board~$43$, Self-hedged~$42$; \cref{tab:s4_descriptives_full}). The median completion time was about $24$ minutes, and participants were compensated through Prolific in line with the platform's recommended rates. All participants provided informed consent before starting and were debriefed afterwards. We compensated the participants with GBP 12.5/hour.

\subsection{Materials and Procedure}
\label{sec:human_procedure}

The study was administered as a web-based grading task linked from Prolific (see Appendix \cref{fig:grader_ui}). After providing consent, participants read instructions describing the collaborative puzzle, the Helper's perspective they would adopt, and the three judgements they would make; they then evaluated a sequence of recorded Worker turns and finally completed a short questionnaire on demographics and prior attitudes toward AI (\cref{tab:s4_survey}). Each rated stimulus showed the flattened Helper--Worker interaction up to a single Worker response, drawn from the first two trials of the benchmark's non-shared-view sessions; for that response the participant reported the correctness likelihood, the apparent confidence, and an intended action (see \cref{sec:human_measures}, Appendix  \cref{fig:grader_ui}). Across conditions we varied only the Worker's message or the visibility of its board (see \cref{tab:conditions_primary}), and each participant remained in a single condition throughout.

\section{Measuring Accuracy and Calibration of Referential Uncertainty}  
\label{sec:study1}  

\begin{table}[t]
\centering
\caption{The three confidence signals for the \emph{same} committed GPT-4.1
placement (image, on-policy; $n=554$ acted turns with ground truth). All signals
score the placed piece, so placement accuracy is shared; we report how well each
signal's confidence tracks correctness. AUROC is discrimination (higher better);
ECE and Brier are calibration (lower better). $T^\star{=}16.5$ is the global
temperature.}
\label{tab:s1_calibration}
\begin{tabular}{lccccc}
\toprule
Confidence signal & Acc.\ (\%) & Mean conf. & AUROC $\uparrow$ & ECE $\downarrow$ & Brier $\downarrow$ \\
\midrule
Raw log-prob                                   & 53.43 & 0.973 & 0.644 & 0.442 & 0.439 \\
Calibrated log-prob ($T^\star{=}16.5$) & 53.43 & 0.531 & 0.623 & 0.157 & 0.271 \\
Elicited belief                                & 53.43 & 0.547 & \textbf{0.648} & \textbf{0.152} & \textbf{0.253} \\
\bottomrule
\end{tabular}
\end{table}  
  
\begin{figure}[t]  
    \centering  
    \includegraphics[width=0.7\textwidth]  
        {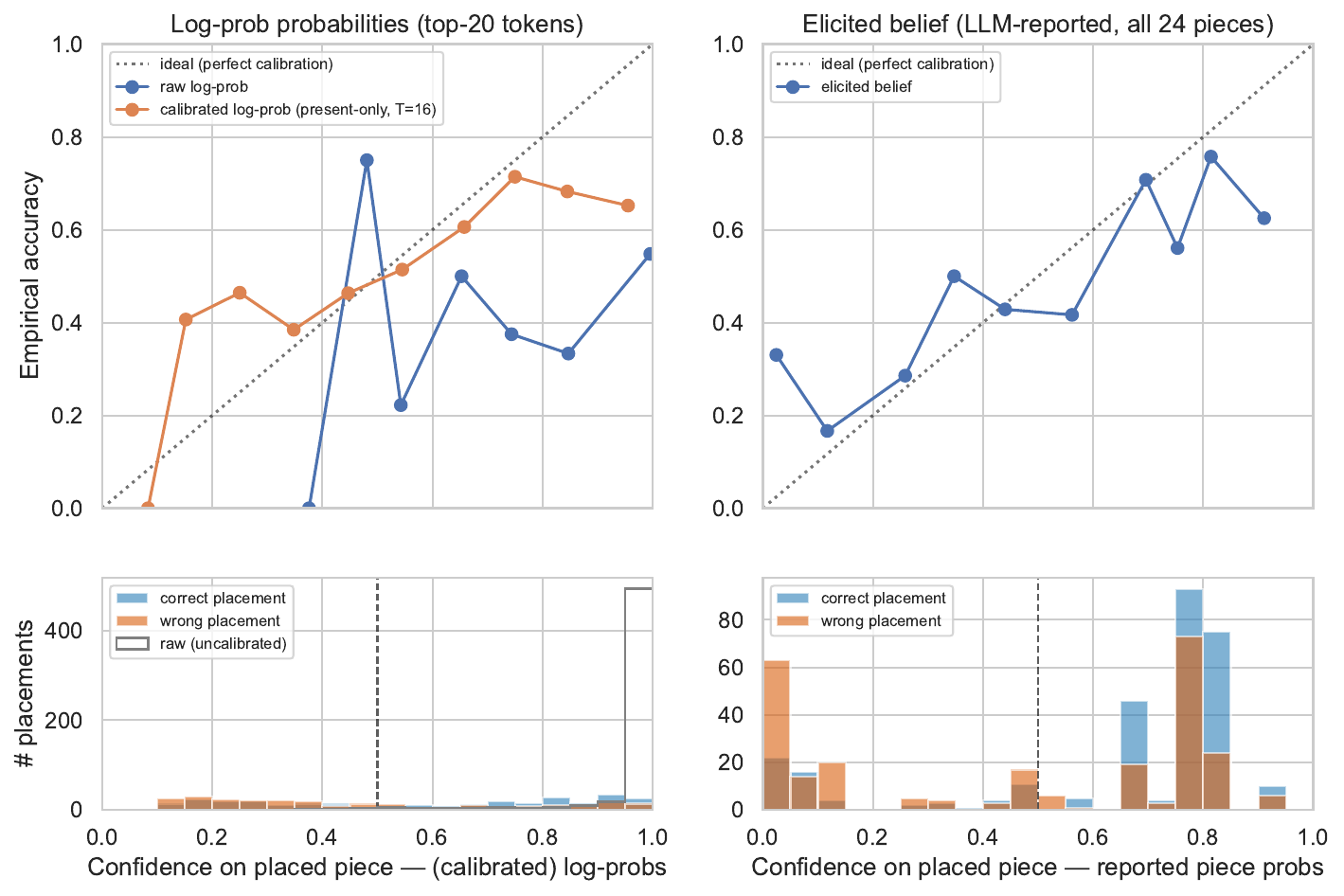}  
    \caption{  
    Reliability curves for the uncertainty signals: (Right) Raw action-token probabilities are  
    over-confident; temperature scaling reduces the gap. (Left) The separately  
    elicited belief provides the strongest combined discrimination and  
    calibration. (Bottom) Distribution of confidence on the placed piece, split by whether 
    the placement was correct or wrong.
    }  
    \label{fig:s1_reliability}  
\end{figure}

We test and measure candidate uncertainty signals that could be used as  deployable \emph{Self-hedged} policy. We first ask whether the model represents, and can express, referential uncertainty. To answer this, we extract three signals that score the same committed placement: the \emph{raw log-prob} (raw action-token log-probabilities of the placed piece-id token), the \emph{calibrated log-prob} (the same log-probabilities after ground-truth-calibrated temperature scaling), and the \emph{elicited belief} (a separately elicited, self-reported confidence distribution over all available pieces). See \cref{sec:measure} for details.

We measure the correctness of the placed piece by the AI worker, given the intended referent (ground truth) and the 
confidence in the referents using the above signals. We measure the following metrics of confidence calibration: AUROC; ECE; Brier score (see Appendix \cref{sec:eval_metrics}).

All three signals discriminate correct from incorrect placements above chance ($\mathrm{AUROC}=0.644$, $0.623$, and $0.648$ for raw, calibrated, and elicited belief, respectively (see \cref{tab:s1_calibration}); $p<10^{-7}$, $p<10^{-6}$, and $p<10^{-9}$). This shows that the model carries  information about whether its referential choice is correct.  
  
The elicited belief is numerically the strongest discriminator, but the pairwise gaps are small and not significant---it sits $\Delta\mathrm{AUROC}=+0.025$ above the calibrated log-probability (95\% CI $[-0.02, 0.08]$, $p=.26$) and $\Delta\mathrm{AUROC}=+0.003$ above the raw log-probability (95\% CI $[-0.03, 0.07]$, $p=.41$)---it is, however, the only signal that pairs this discrimination with good calibration (overconfidence $+0.01$, 95\% CI $[-0.03, 0.05]$; $\mathrm{ECE}=0.15$), and the only one available for the models that expose no token log-probabilities, see \cref{fig:s1_reliability}.
  
The raw action-token probabilities are severely overconfident (mean confidence $0.97$ vs.\ $53\%$ accuracy; overconfidence $+0.44$, 95\% CI $[0.40, 0.48]$). Temperature scaling removes this  ($T^\star=16.5$; overconfidence $\approx0$) but gives the numerically lowest discrimination ($\mathrm{AUROC}=0.623$)\footnote{Calibrated confidence is not a monotone transform of the raw token probability (Kendall $\tau=0.52$)---it renormalizes the (temperature-scaled) placed-piece probability over the 24 pieces, so its per-turn denominator, and hence the ranking used by AUROC, depends on how many pieces appear in the top-20 that turn.}. Still, being better calibrated is not the same as being useful in this case — the calibrated log‑prob's correct and wrong distributions are nearly indistinguishable, while the elicited belief additionally separates them.

For the remaining analyses, we use the separately elicited belief distribution. It combines better calibration than raw action-token probabilities with stronger discrimination than either raw or temperature-scaled probabilities. It also provides a consistent uncertainty measure across models, we do not require access to token-level log-probabilities, which are not exposed by all model APIs and are unavailable for GPT-5 and GPT-5.5 in our setup.

Next, we perform a systematic analysis on the sensitivity of the elicited confidences over specific levels of uncertainty about the referents. To compare models without collecting new live interactions, we replay individual decision points using a single prompt. See Appendix \cref{sec:models}.

\section{Robustness Analysis - Task Difficulty and Referential Uncertainty} \label{sec:study2}  

We are testing how sensitive the elicited uncertainties from the models are to different levels of vagueness in both Helper instructions and board context. We test: (a) influence of the modality of the context (images vs. textual description); (b) instruction specificity (unambiguous to vague); (c) board confusability (no alternative matching target pieces vs. a group of ambiguous pieces to the target).

\subsection{Context Modality}  
\label{sec:s2_modality}  
In the original puzzle task~\cite{poelitz2026benchmarkassesscommonground}, the puzzle context was added to the model context as byte-64 encoded png-image (needed visual encoding which might additional challenges for the model to express and act on uncertainty). We reproduced the original study using the single prompts with collated conversation (as described above) using an explicit textual description of the puzzle boards and pieces instead of the images (see Appendix \cref{fig:modality_example}). We compare this with the image encoded results across different models using the original dialogues.  

Replacing the rendered board and pieces with an explicit textual description improves accuracy for the two earlier models: on turns  
where the model committed a placement under both modalities, GPT-4.1 rises from $52.9\%$ to $66.0\%$, and GPT-5 from $58.4\%$ to $71.7\%$  (both $p<0.001$). GPT-5.5 is more modality-robust, with no significant difference ($66.7\%$ vs.\ $70.1\%$, $p=0.16$). 

The models differ in how often they act i.e, place a piece at all. GPT-4.1 and GPT-5.5 place a piece on the large majority of instructed turns under both modalities ($\sim85$--$90\%$), whereas GPT-5 produces a \textbf{no-action} far more often, and even more under text ($23\%$ of turns for images vs.\ $34\%$ for  text). We report the no-action rate here only descriptively and defer the analysis of \emph{how} models ask for clarification to \cref{sec:study3}.  
  
The differences on the elicited belief-entropy differences are small and do not track the accuracy gain. GPT-4.1 is slightly more uncertain under images than text ($1.74$ vs. $1.56$ nats; Wilcoxon $p<0.001$, rank-biserial $r=0.36$), consistent with text being the easier encoding; GPT-5, despite being \emph{more} accurate under text, is marginally \emph{more} uncertain here ($0.78$ vs.\ $0.84$ nats; $p<0.001$, $r=-0.23$); and GPT-5.5 shows no difference ($p=0.62$). Hence, the modality that yields the better placements is not reliably the one with lower elicited uncertainty. (See also Appendix \cref{fig:s2b_modality}).
  
\subsection{Instruction Specificity}  
\label{sec:s2_instruction}  

\begin{figure}[t]  
    \centering  
    \includegraphics[width=1\textwidth]  
        {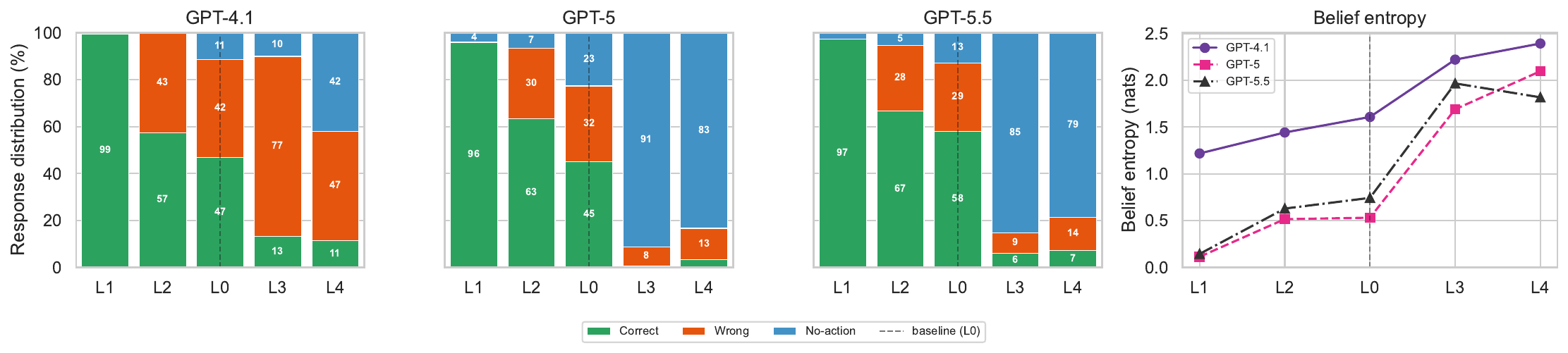}  
    \caption{Instruction specificity (image modality) results per model. (left) stacked outcome composition of Worker response (correct piece / wrong piece / no-action) at each specificity level from overspecified (L1) to vague (L4), with the original utterance (L0) placed between L2 and L3 as the baseline (dashed); the right panel shows mean belief entropy. Accuracy falls and belief entropy rises with vagueness for all models.}
    \label{fig:s2a_levels}  
\end{figure} 

Next, we study the effects on the elicited referential uncertainty on how specific the user instructions (Helper) are. For each Helper turn which resulted in the Worker to place a piece in the puzzle task benchmark~\cite{poelitz2026benchmarkassesscommonground}, we rephrase what the Helper said with a specific level of specificity from overspecified (L1) to vague (L4), with (L0) the original Helper instruction (See Appendix \cref{tab:levels} for more details and \cref{tab:instr_templates} for the templates used to generate the specific instruction.). 
 
Across all three models and both modalities (see \cref{fig:s2a_levels} for images and Appendix \cref{fig:s2a_levels_text} for text), referential accuracy falls and elicited belief entropy rises monotonically as the instruction becomes vaguer; every trend is significant at $p<0.001$ (see also Appendix \cref{tab:s2a_trend}).  Coverage---the share of instructed turns on which the model commits a placement---also decreases significantly, i.e.\ the models increasingly produce a no-action rather than guess; we analyse this behaviour in \cref{sec:study3}.  
  
For GPT-4.1, each step toward vagueness costs about $21$ points of accuracy ($b=-21.2$, 95\% CI $[-22.9,-19.6]$,  
$t(24)=-24.9$, $p<0.001$) and adds about $0.31$ nats of belief entropy ($b=+0.313$, 95\% CI $[0.275,0.351]$, $t(24)=16.0$, $p<0.001$). The newer models register the same vagueness more through \emph{withholding}: their coverage drops roughly $24$ points per step (vs.\ $9$ for GPT-4.1), and their entropy rises about $0.5$ nats per step (vs.\ $0.31$).  

\subsection{Board Confusability}  
\label{sec:s2_clusters}  
\begin{figure}[t]  
    \centering  
    \includegraphics[width=1\textwidth]  
        {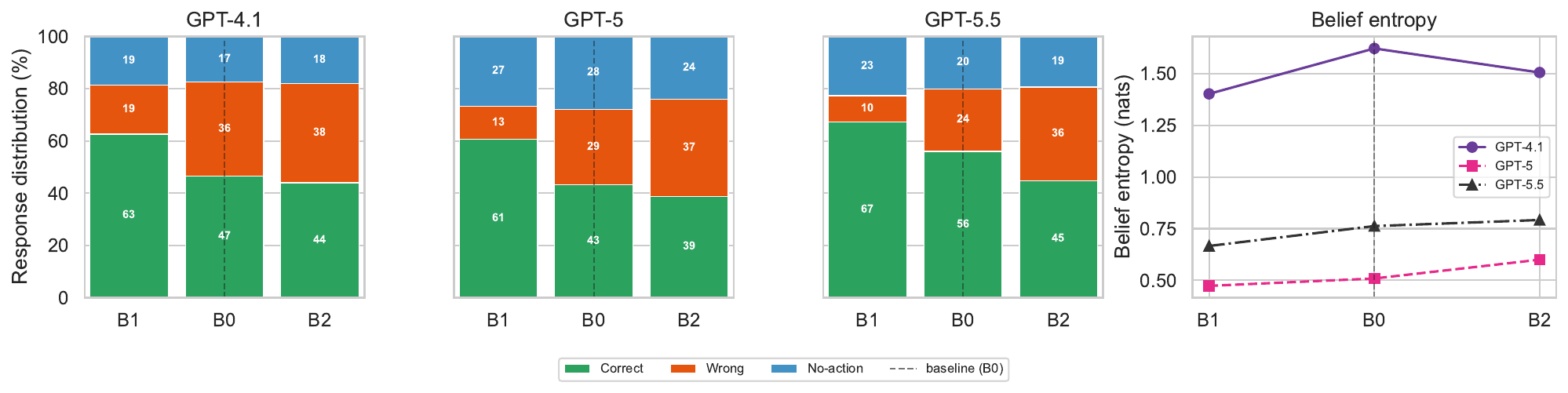}  
    \caption{Board confusability (image modality) results per model. (left) stacked outcome
composition of Worker responses (correct piece / wrong piece / no-action) and (right) mean belief entropy on a distinct board (B1), the original board (B0), and a confusable board (B2), in order of increasing confusability. Accuracy falls sharply while no-action and belief entropy barely move.}  
    \label{fig:s2c_boards}  
\end{figure}  

Finally, we test the effect of confusability of the puzzle pieces with the target. We design two new board configurations for each target piece: (B1) only distinct pieces to the target pieces on the board; (B2) a set of clear similar pieces to the target on the board; and (B0) the original board. See Appendix \cref{fig:board_confusability} for an illustration. 

Confusability sharply reduces accuracy---by $12$ to $25$ points per step from the distinct to the confusable board for every model and modality (all $p<0.001$; See Appendix \cref{tab:s2c_trend})---a decline comparable  
to instruction vagueness. Unlike vagueness, this is barely present in the model's expressed uncertainty. Coverage shows no significant trend for any model (the sole exception being a small decrease for GPT-5.5 in with text modality; see Appendix \cref{fig:s2c_boards_text}), so confusability does not prompt the models to withhold a placement, in contrast to the steep coverage drop under vagueness. Belief entropy does rise, but only by $0.05$ to $0.19$ nats per step---roughly a fifth of the vagueness effect---though significantly for all three models. 

\begin{table}[t]\centering\small
\begin{tabular}{lcccc}
\toprule
 & argmax switches & in cluster & $\Delta$peak on switch & $\Delta$peak when kept \\
\midrule
image & 38\% & 79\% & $-0.07$ & $-0.06$ \\
text  & 46\% & 58\% & $-0.12$ & $-0.16$ \\
\bottomrule
\end{tabular}
\caption{The belief peak relocates onto the same cluster. The argmax switches on
$38\%$/$46\%$ of turns (image/text) between the distinct (B1) and confusable (B2) boards;
of these, $79\%$/$58\%$ land inside the confusable cluster (vs.\ ${\sim}15\%$ chance,
binomial $p<10^{-45}$). The peak-height drop is the same whether the argmax switched or was
kept ($\Delta_{\text{switch}}\approx\Delta_{\text{kept}}$), so switching to a look-alike
carries no extra loss of confidence.}
\label{tab:2c_conf_switches}
\end{table}

We observed that errors climb with confusability while belief entropy barely moves. We therefore ask where the probability mass goes when similar referents are added to the board. Rather than spreading its belief across the confusable set of similar referents---which would raise entropy---the model relocates its \emph{peak} onto a single competing piece from that set.  
  
Comparing the confusable board (B2) with the distinct board (B1) on the same turns, the belief argmax switches on $38\%$ of turns for images and $46\%$ for text; of these switches, $79\%$ (image) and $58\%$ (text) land on a same-family cluster, far above the $15\%$ rate expected by chance (binomial $p<10^{-45}$; \cref{tab:2c_conf_switches}). The peak itself stays tall: its height falls only modestly (more so in text) and by a comparable amount whether the argmax moved or was kept  ($\Delta_{\mathrm{switch}}\approx\Delta_{\mathrm{kept}}$), so the model commits to the wrong look-alike almost as confidently as to the target.  
  
Because the placement follows the argmax, this one-slot shift turns a correct placement into a wrong one while the belief remains peaked---explaining why accuracy falls sharply yet entropy hardly rises. 
 
\section{Externalizing of Referential Uncertainty}  
\label{sec:study3}  

In the previous experiments, we saw that the models do not always perform an action and that there is a trend towards no-action with increasing level of ambiguity in the user instructions. In contrast, we did not observe the same trend on the level of ambiguity in the amount the referents. Next, we inspect if and how these no-actions are expressing uncertainty and whether they ask for clarification. We check if they are generic hedges or specific questions targets on the piece with ambiguity. 
  
\subsection{Clarification Behaviour}  
\label{sec:s3_clarify}  

\begin{figure}[t]  
    \centering  
    \includegraphics[width=0.7\textwidth]  
        {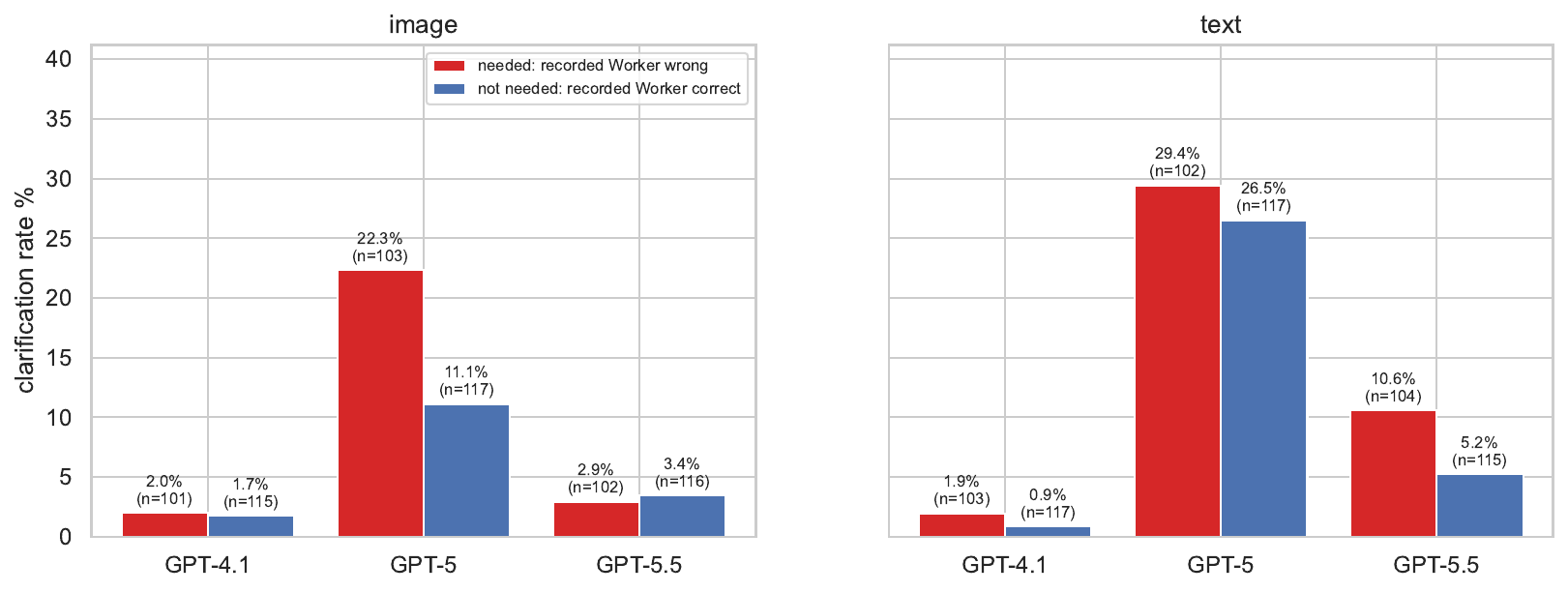}  
    \caption{  
    Clarification rates on turns with incorrect piece selected (could benefit from clarification) and correct select piece (would not benefit from clarification) across models model.
    }  
    \label{fig:s3a_clarification}  
\end{figure}  

First, we measure how often the models ask a clarification question at each turn with a Helper instruction to place a piece. Using the ground truth pieces, we also compare to the recorded Worker's placements and check if a question here would have possibly helped in the original study. Hence, we measure whether the model would have made a better decision and rather ask than place. (GPT4.1 places a piece almost all the time in the original study.) To measure if a model asked a question, we first filter for all responses which are not an action to place a piece. Further, we use GPT5.5 to label the response as clarification question (see Appendix for the prompts).

We observe different numbers of clarification across models and modality (see \cref{fig:s3a_clarification}) GPT-5 asks by far the most clarifications. It asks for clarification in $16.7\%$ of image turns and $23.8\%$ of text turns, versus $3.5\%$/$1.6\%$ for GPT-4.1 and $6.2\%$/$8.8\%$ for GPT-5.5 (See Appendix \cref{tab:s3a_clarify}). Almost all are explicit questions that name the confusable piece rather than generic hedges (hedging is $<1\%$ for every model).  
  
Taking the recorded Worker's placement as the proxy for when a question would have helped (see Appendix \cref{tab:s3a_targeting}), GPT-5's clarifications in image pieces concentrate significantly on erroneous turns---$22.3\%$ where the Worker placed the wrong piece versus $11.1\%$ where it was correct ($p=0.020$). On text pieces, however, the clarification rate is not significantly different ($29.4\%$ vs.\ $26.5\%$, $p=0.37$). GPT-4.1 and GPT-5.5 ask clarifications too rarely to show reliable targeting in either modality ($p\geq 0.11$).

\subsection{Natural-Turn Error Prediction and Hedge Triggering}  
\label{sec:s3_trigger}  

\begin{figure}[t]  
    \centering  
    \includegraphics[width=0.7\textwidth]  
        {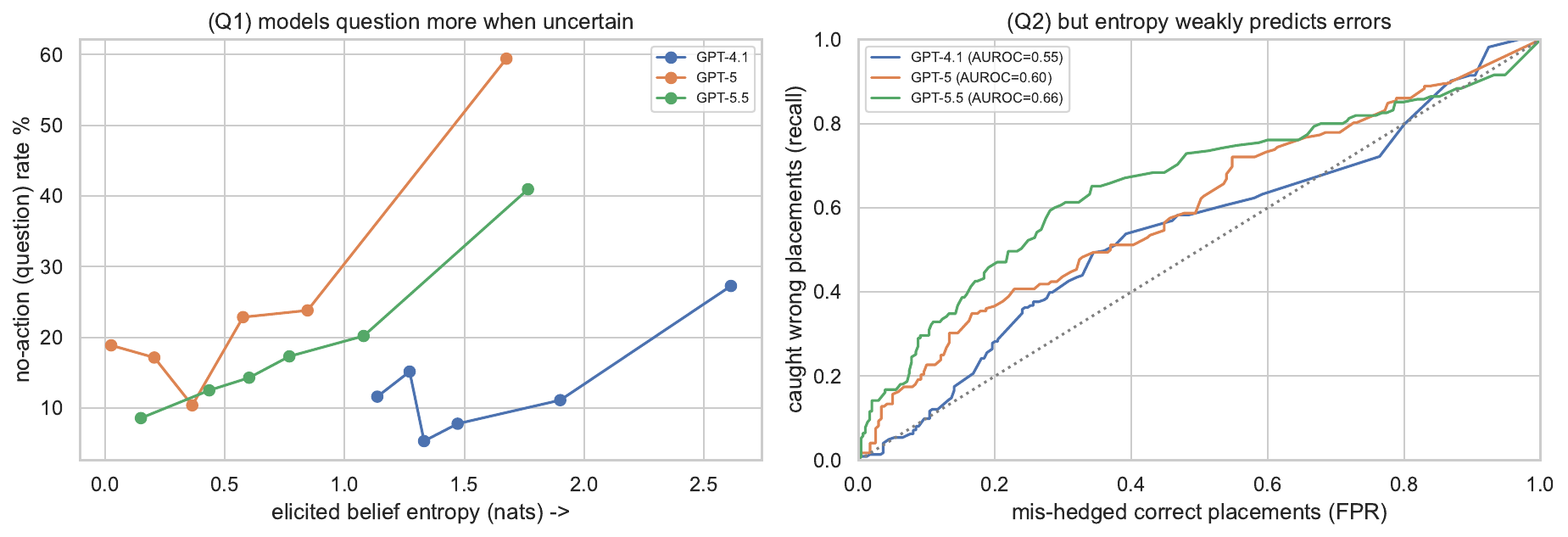}  
    \caption{Elicited uncertainty as an externalization signal (image modality).
(Left) the no-action (question) rate rises with the model's own elicited belief entropy for all
three models---they question more when their belief is more spread out. (Right) ROC of the rule
``hedge if belief entropy $\geq\tau$'' for predicting the model's own wrong placements (scored vs.\
ground truth piece); all three curves sit close to the diagonal (AUROC $0.55$--$0.66$), so belief
entropy is only a weak error predictor. Models externalize uncertainty in proportion to a signal that is
itself too weak to cleanly trigger a hedge.}  
    \label{fig:s3_underasking}  
\end{figure}

The results above show two things: First, models do externalize uncertainty on the turns where they decide to act (see left \cref{fig:s3_underasking}): the no-action rate rises with elicited belief entropy, and entropy is reliably higher on question turns than on placement turns (point-biserial $r=0.18$--$0.39$, all $p<10^{-4}$;  
see Appendix \cref{tab:s3_trigger}).  
  
Second, that signal is a weak error predictor (see right \cref{fig:s3_underasking}). On turns where the model does act, belief entropy separates its wrong from its correct placements only weakly (AUROC $0.55$--$0.66$). A rule that hedges whenever entropy exceeds its median would catch $57$--$68\%$ of wrong placements but mis-hedge $41$--$48\%$ of correct ones, with precision barely above the base error rate. Elicited confidence is thus a genuine but weak externalization signal: models ask in proportion to it, yet it is too coarse to serve as a reliable hedge trigger.  

What ultimately matters, however, is the \emph{human Helper's} perception. We therefore turn to the human Helper and ask whether communicating this uncertainty---together with more informative piece descriptions---helps a partner tell correct from incorrect placements and act on them, and whether the model's own weak self-signal is sufficient or an oracle-targeted hedge is required.




\section{Human Perception on Confidence and Correctness}  
\label{sec:study_results}  
We conducted a between-subject study where we presented participants an interaction history for the puzzle tasks up to a current turn with a Worker's response for a Helper's instruction to place a piece. 

We restrict the study to original puzzle tasks which did not share the puzzle board and hence the Helper did not see the Worker's actions. This was more challenging than the shared view since the Helper (could/did) not correct the Worker. Further, we only grade the first two puzzles. On these puzzle tasks the accuracy on the pieces was only $36\%$, compared to the $53\%$ on all tasks. See \cref{tab:belief_study_turns} for comparison.

\begin{table}[t]
\centering\small
\caption{Comparison of GPT-4.1's elicited referential belief between the full set of turns and the subset used in this user study.}
\label{tab:belief_study_turns}
\begin{tabular}{lccccc}
\toprule
Turn set & $n$ & Acc.\ (\%) & AUROC $\uparrow$ & ECE $\downarrow$ & Brier $\downarrow$ \\
\midrule
All turns & $554$ & $53.4$ & $0.648$ & $0.152$ & $0.253$ \\
Study turns & $228$ & $36.4$ & $0.5$ & $0.207$ & $0.267$ \\
\bottomrule
\end{tabular}
\end{table}


For the \emph{Self-Hedged} condition, we use a threshold $\tau$ such that $H>\tau$, with $\tau{=}1.34$~nats. We fix $\tau$ by \emph{rate matching} on GPT-4.1's original turns: $\tau$ is set so that the hedge-firing rate ($51\%$) matches the empirical wrong-placement rate ($52\%$), letting the deployable self-hedge placing the hedge on about as many turns as the oracle \emph{without} using ground truth---but, because belief entropy is only a weak error signal (\cref{sec:s3_trigger}), at much lower precision (its flagged-turn counts are in \cref{tab:s4_mechanism}).


\subsection{Within Condition Results}

\begin{table}[t]
\centering
\small
\caption{Human study results per-condition (participant means; $N=210$: $168$ in the four primary conditions plus $42$ in the exploratory Self-hedged condition). More detailed and hedged messages lower the perceived correctness and---only under hedging---apparent confidence; rating and action \emph{discrimination} (correct$-$wrong) rise in the Hedged condition, compared to the Generic condition, with similar results as in the Visible condition;  Self-hedged condition does not show this similarly.}
\label{tab:s4_results}
\begin{tabular}{lccccc}
\toprule
& Generic & Described & Hedged & Visible board & Self-hedged \\
\midrule
Perceived correctness (0--100)          & 77.6 & 67.3 & 56.1 & 53.7 & 61.5 \\
Apparent confidence (0--100)            & 85.5 & 85.7 & 58.5 & 83.3 & 56.6 \\
Rating discrimination (corr.$-$wrong)   & $-0.1$ & 16.8 & 27.7 & 26.6 & 8.8 \\
\midrule
Accept correct moves (\%)               & 79.0 & 78.6 & 73.7 & 64.8 & 60.1 \\
Accept wrong moves (\%)                 & 78.3 & 57.1 & 36.4 & 30.1 & 50.2 \\
Action discrimination (pp)              & 0.7 & 21.5 & 37.3 & 34.8 & 10.0 \\
\bottomrule
\end{tabular}
\end{table}

As the Worker was correct on only $\approx 36\%$ pieces, a well-calibrated Helper should discriminate sharply and often decline to proceed. In the \emph{Generic} condition, we observe neither. Graders rate nearly every placement as likely correct ($77.6/100$) and highly confident ($85.5$), accept $78.3\%$ of \emph{wrong} moves, and shows no discrimination (rating gap $-0.1$; action gap $0.7$~pp). See \cref{tab:s4_results} for more details. 
  
A precise description (in condition \emph{Described}) lowers perceived correctness to $67.3$ and lowers wrong-move acceptance by $21.2$~pp (to $57.1\%$), with an increase in rating and action discrimination to $16.8$ and $21.5$~pp. On the other hand, the apparent confidence does not change ($85.7$). Hence, the description can convey what was placed, not how sure the Worker is. Only the two hedged conditions reduce the apparent confidence (in condition \emph{Hedged} to $58.5$; in condition \emph{Self-hedged} to $56.6$).  
  
The \emph{Hedged} condition is the strongest, excluding the shared-view on the board, reducing wrong-move acceptance to $36.4\%$ and increasing rating and action discrimination to $27.7/37.3$~pp—matching the \emph{Visible board} condition results ($26.6/34.8$~pp; $30.1\%$ wrong accepted), despite the human never seeing the board—at only a modest cost on correct moves (accept-correct $73.7\%$ vs.\ $78.6\%$). The \emph{Self-hedged} condition shows the same low-confidence ($56.6$) but the graders still accept $50.2\%$ of wrong moves and discriminate weakly ($8.8/10.0$~pp), close to the Generic/Described floor. (See Appendix \cref{tab:s4_descriptives_full} for individual results.)  

\subsection{Between Condition Results}

\begin{table}[t]
\centering
\small
\caption{Study results on pairwise contrasts: participant-mean differences $\Delta$ (positive $=$ first
condition higher). Accept-wrong and action discrimination are in percentage points; correctness,
apparent confidence and rating discrimination are in rating points. Holm is corrected within each
family over the $10$ behavioural decision outcomes ($60$ primary / $40$ secondary tests); the
questionnaire and validity/randomisation checks are corrected/reported separately. \textbf{Bold}
$=$ survives the global (within-family) Holm correction; $^{*}=$ significant under within-outcome
Holm but not global; unmarked $=$ n.s. $95\%$ bootstrap CIs for the headline effects are in the text.}
\label{tab:s4_pairwise}
\begin{tabular}{lccccc}
\toprule
Contrast & Correctness & Perceived conf. & Rating discr. & Accept-wrong (pp) & Action discr. (pp) \\
\midrule
\multicolumn{6}{l}{\emph{Primary family}}\\
Described $-$ Generic     & $-10.3^{*}$ & $0.2$ & $\mathbf{16.9}$ & $\mathbf{-21.2}$ & $\mathbf{20.7}$ \\
Hedged $-$ Described      & $-11.2^{*}$ & $\mathbf{-27.3}$ & $10.9$ & $\mathbf{-20.7}$ & $15.9^{*}$ \\
Hedged $-$ Visible board  & $2.4$ & $\mathbf{-24.9}$ & $1.1$ & $6.3$ & $2.6$ \\
Visible board $-$ Generic & $\mathbf{-23.9}$ & $-2.2$ & $\mathbf{26.8}$ & $\mathbf{-48.3}$ & $\mathbf{34.1}$ \\
\midrule
\multicolumn{6}{l}{\emph{Secondary family (Self-hedged)}}\\
Self-hedged $-$ Hedged    & $5.3$ & $-1.9$ & $\mathbf{-19.0}$ & $13.8^{*}$ & $\mathbf{-27.3}$ \\
Self-hedged $-$ Described & $-5.8$ & $\mathbf{-29.1}$ & $-8.1$ & $-6.9$ & $-11.5$ \\
\bottomrule
\end{tabular}
\end{table}

We report participant-mean contrasts with the \emph{Generic} condition as reference in \cref{tab:s4_pairwise}. 
Relative to the \emph{Generic} condition, in the \emph{Described} condition, rating discrimination increased by $16.9$ points (95\%~CI [$8.7$, $25.4$]) and action discrimination by $20.7$~pp. Wrong-move acceptance was reduced by $21.2$~pp (about $27\%$). All three effects are statistically significant; on the other hand the confidence perception is unchanged ($+0.2$, n.s.). 
  
Comparing the \emph{Hedged} condition with the \emph{Described} condition, perceived confidence reduces by $27.3$ points (95\%~CI [$-33.1$, $-21.3$], $p<.001$) and wrong-move acceptance reduces by $20.7$~pp ($p=.049$)—both effects are statistically significant, but only the improvement in action discrimination ($+15.9$~pp) is significant within outcome, while the rating-discrimination gain ($+10.9$ points) is not significant.  
  
The \emph{Hedged} condition matches the \emph{Visible-board} condition on decision quality but not on perceived confidence. The \emph{Hedged} condition and the  \emph{Visible board} condition are indistinguishable on every decision-quality outcome (correctness: $p=.56$; rating discrimination: $p=.86$; accept-wrong: $p=.31$; action discrimination: $p=.75$), yet we observe $24.9$ fewer points of perceived confidence (95\%~CI [$-30.7$, $-18.9$], $p<.001$ globally) in the \emph{Hedged} condition. Board access itself (\emph{Visible board} versus \emph{Generic}) improves all four decision-quality outcomes (all statistically significant) but leaves apparent confidence unchanged ($-2.2$, n.s.). 

The apparent confidence in the \emph{Self-hedged} condition is comparable to the \emph{Hedged} condition ($-1.9$, $p=.51$) and is $29.1$ points lower than in the \emph{Described} condition ($p<.001$). Relative to the \emph{Hedged} condition, in the \emph{Self-hedged} condition we observe a $19.0$ points lower rating discrimination and $27.3$~pp lower action discrimination (both statistically significant) and a $13.8$~pp increase in acceptance of wrong moves. Relative to the \emph{Described} condition, the discrimination is not a statistically significant difference (rating: $-8.1$, $p=.09$; action: $-11.5$~pp, $p=.14$).

\subsection{Hedging Mechanism}
\begin{table}[t]
\centering\small
\caption{Ratings by turn type for the two hedging designs,
each scored against \emph{Described} on the turns \emph{that design flags}. Hedged flags by the oracle
(objectively wrong turns; $36$ wrong / $0$ false alarms, precision $100\%$); Self-hedged flags from the
Worker's own belief entropy ($29$ wrong / $13$ false alarms, precision $69\%$). Turn type crosses whether
the design flagged the turn with whether the Worker was actually correct; acceptance is in percentage
points, apparent confidence and correctness rating in $0$--$100$ points. \emph{Described} appears in both
blocks because each restricts to that design's flagged-turn set, and the oracle's false-alarm row is empty
by construction. Confirmatory contrasts are participant-level permutation tests, Holm-corrected within
family (the oracle contrast also passes an exact dialogue-level sign-flip test); the isolated
false-alarm sub-row is descriptive. Effect sizes, $95\%$ CIs, and $p$-values are reported in the text.}
\label{tab:s4_mechanism}
\begin{tabular}{l cc c cc}
\toprule
 & \multicolumn{2}{c}{\textbf{Oracle targeting (Hedged)}} & & \multicolumn{2}{c}{\textbf{Self targeting (Self-hedged)}}\\
\cmidrule(lr){2-3}\cmidrule(lr){5-6}
Turn type $\times$ outcome & Described & Hedged & & Described & Self-hedged\\
\midrule
\multicolumn{6}{l}{\emph{Flagged wrong} (caught error)}\\
\quad Apparent confidence & 86.5 & 36.3 & & 86.3 & 33.6\\
\quad Correctness rating  & 63.4 & 46.1 & & 61.5 & 51.9\\
\quad Acceptance (\%)      & 59   & 34   & & 57   & 41\\
\midrule
\multicolumn{6}{l}{\emph{Flagged correct} (false alarm)}\\
\quad Apparent confidence & --   & --   & & 87.5 & 35.2\\
\quad Correctness rating  & --   & --   & & 81.9 & 59.3\\
\quad Acceptance (\%)      & --   & --   & & 87   & 49\\
\midrule
\multicolumn{6}{l}{\emph{Unflagged wrong} (missed error)}\\
\quad Apparent confidence & 85.0 & 69.9 & & 86.5 & 78.8\\
\quad Correctness rating  & 55.9 & 41.5 & & 61.1 & 67.3\\
\quad Acceptance (\%)      & 51   & 37   & & 56   & 62\\
\midrule
\multicolumn{6}{l}{\emph{Unflagged correct} (left alone)}\\
\quad Apparent confidence & 88.6 & 77.9 & & 90.0 & 81.8\\
\quad Correctness rating  & 78.3 & 73.4 & & 75.1 & 73.4\\
\quad Acceptance (\%)      & 79   & 74   & & 69   & 72\\
\bottomrule
\end{tabular}
\end{table}

We report comparison results of the two hedging conditions side by side, each scored against the \emph{Described} condition on the turns that are hedged in \cref{tab:s4_mechanism}.

In the \emph{Hedged} condition, (oracle flags) we put hedge exactly on wrong turns and acceptance falls $59\%\!\to\!34\%$ ($-24.8$~pp, Holm $p{=}.014$; resulting in a $50$-point decrease in perceived confidence and $17$-point decrease in correctness estimation (all statistically significant). The correct turns are not effected ($79\%\!\to\!74\%$, n.s.). 
Note, the false-alarm row is empty as the oracle flags $36$ wrong and $0$ correct turns. 

In the \emph{Self-hedged} condition, we use the above described 
belief-entropy to hedge. At $69\%$ precision (see above) the hedges mirror this: on the errors it catches, the acceptance discount is weaker and not significant ($57\%$ compared to $41\%$, $-15.8$~pp, $p{=}.22$) and on the correct turns it wrongly flags, acceptance reduces from $87\%\!\to\!49\%$ ($-37$~pp). On the the error the hedge misses acceptance increase ($56\%\!\to\!62\%$). 

None of the difference between the \emph{Described} and the \emph{Self-hedged} conditions are significant---acceptance falls $-15.8$~pp ($p{=}.22$) on the errors the self-hedge flags and $-18.4$~pp ($p{=}.09$) on correct turns overall (the flagged-correct false alarms alone show a descriptive $-37$~pp drop). So the misplaced hedge reduces the confidence estimation (desired) and the correctness estimation (undesired). Hence, its discrimination is similar to the \emph{Described} condition, unlike the \emph{Hedged} condition.

\section{Conclusion}
\label{sec:conclusion}

We studied \emph{referential uncertainty}---uncertainty over which object a description refers to---in a collaborative puzzle task, decomposing it into four capabilities: the Worker \emph{identifying} and \emph{communicating} this uncertainty, and the Helper \emph{recognizing} and \emph{acting} on it.  

On the representation side, the models are capable. A separately elicited belief distribution over candidate pieces is better calibrated and more discriminative of correct from incorrect placements than raw action-token probabilities, which are severely over-confident (\cref{sec:study1}, \cref{tab:s1_calibration}). This elicited uncertainty tracks controlled task difficulty: it increases as instructions become more vague (\cref{sec:s2_instruction}), while context modality (visual vs.\ textual) has only small, model-dependent effects (\cref{sec:s2_modality}). It is not uniformly sensitive, however: confusable alternatives sharply raise error while barely moving the elicited uncertainty (\cref{sec:s2_clusters}), because the belief peak relocates onto a single lookalike rather than spreading out.  

On the communication side, they fall short. The models seldom ask for clarification or hedge, even when their error rate would warrant it, instead echoing the Helper's brevity (\cref{sec:s3_clarify}). Their own belief entropy is a genuine but weak per-turn error predictor (AUROC $0.55$--$0.66$; \cref{sec:s3_trigger}), so a simple ``hedge when entropy is high'' rule is not yet a deployable policy.  

These signals are what human partners need. In a controlled study, participants given only the generic Worker message accept most wrong placements and cannot tell correct from incorrect; precise piece descriptions and, especially, well-targeted hedges restore discrimination and roughly halve wrong-move acceptance, matching a fully visible board without ever revealing it (\cref{tab:s4_results}). The benefit, though, depends on \emph{which} turns are hedged: a deployable self-hedge derived from the model's own belief entropy inherits that signal's weakness and can do more harm than good, also drawing attention to correctly placed pieces.  

The bottleneck is therefore not uncertainty estimation alone but the quality and targeting of the uncertainty that is \emph{externalized}. Future work should therefore pursue two complementary directions: fine-tuning to better calibrate the elicited belief~\cite{lin2022tmlr-teaching, liu2026reinforcement}, and techniques that adapt \emph{when} to externalize it~\cite{kumaran2026causal}---drawing on more than raw entropy, such as the full belief distribution, candidate margins, instruction ambiguity, and context confusability. Throughout, correctness must be scored against the intended referent rather than the worker's observed placement.  

\bibliographystyle{ACM-Reference-Format}
\bibliography{references}  

\clearpage
\appendix
\section{Appendix}
\subsection{Dataset Summary}
\begin{table}[htbp]
\centering\small
\caption{Statistics of the source Helper--Worker interactions from the
collaborative puzzle benchmark~\cite{poelitz2026benchmarkassesscommonground}
that our analyses draw on (curated, referent-annotated dialogues).}
\label{tab:dataset_stats}
\begin{tabular}{@{}lr@{}}
\toprule
\textbf{Property} & \textbf{Value} \\
\midrule
\multicolumn{2}{@{}l}{\emph{Source dialogues}}\\
\quad Dialogues (session\,$\times$\,puzzle)      & $206$ \\
\quad Sessions (Helper--Worker pairs)            & $44$ \\
\quad View (shared / non-shared)                 & $105$ / $101$ \\
\quad Puzzles per session                        & $5$ (P1--P5, ${\sim}41$ each) \\
\quad Target size (blocks)                        & $4$ \\
\quad Completed (pair agreed solved)             & $164$ / $206$ \\
\midrule
\multicolumn{2}{@{}l}{\emph{Messages}}\\
\quad Total (Helper / Worker / system)           & $3{,}651$ ($1{,}727$ / $1{,}718$ / $206$) \\
\quad Per dialogue (mean / median / range)       & $17.7$ / $14$ / $2$--$68$ \\
\bottomrule
\end{tabular}
\end{table}

\FloatBarrier
\subsection{Ground-Truth Scoring}  
\label{sec:ground_truth}  
We manually annotate the intended referent ($y_t^{\texttt{gt}}$) for each evaluated placement turn in the source dataset. A Worker's observed placement is not used as the correctness target. The Worker may itself have selected an incorrect piece, so agreement with its action does not establish agreement with the Helper's intended referent.  
  
For a turn $t$ on which the Worker selects piece $\widehat{y}_t$, referential correctness is defined as  
\begin{equation}  
    \label{eq:referential_correctness}  
    z_t  
    =  
    \mathbb{I}\!\left[  
        \widehat{y}_t =  
        y_t^{\texttt{gt}}  
    \right].  
\end{equation}    
  
We distinguish incorrect selections from turns on which the model makes no placement. Accuracy on committed placements will be reported together with \emph{coverage}, the proportion of eligible instruction turns on which the model places a piece. A no-action response is not automatically treated as a clarification request.  

\subsection{Rating and Action Discrimination}
\label{sec:discriminations}
Let $r_{jt}$ denote participant $j$'s correctness rating for turn $t$, let $a_{jt}=1$ indicate acceptance, and let $z_t=1$ indicate that the selected piece matches the intended referent. We define \emph{rating discrimination} as  
\begin{equation}  
    D^{\mathrm{rating}}_j  
    =  
    \operatorname{mean}(r_{jt}\mid z_t=1)  
    -  
    \operatorname{mean}(r_{jt}\mid z_t=0),  
\end{equation}  
and \emph{action discrimination} as  
\begin{equation}  
    D^{\mathrm{action}}_j  
    =  
    P_j(a_{jt}=1\mid z_t=1)  
    -  
    P_j(a_{jt}=1\mid z_t=0).  
\end{equation}  

Positive values indicate greater differentiation between correct and incorrect selections. We also report acceptance rates for correct and incorrect selections separately, since an overall reduction in acceptance does not indicate improved discrimination. Action discrimination is reported in percentage points.
  
\FloatBarrier
\subsection{Measuring Referential Uncertainty}  
\label{sec:measure}  
  
We operationalize referential uncertainty as a probability distribution over the candidate pieces available to the Worker. We compare three signals: raw action-token probabilities, temperature-scaled action-token probabilities, and a separately elicited belief distribution. These are observable model outputs, not direct measurements of an internal belief state.  

\FloatBarrier
\paragraph{Raw action-token probabilities.}  
For GPT-4.1, we obtain the top-$K$ token log-probabilities ($K=20$) at the piece-identifier position in the generated action, such as \texttt{PLACE piece \#13 at (1,2)}. We map tokens to piece identifiers, sum probabilities for tokens referring to the same piece, and normalize over the represented pieces. This yields a distribution conditional on the candidate identifiers recovered from the returned token probabilities, rather than a complete distribution over all available pieces.  

\FloatBarrier
\paragraph{Temperature-scaled probabilities.}  
We apply a single global temperature to the candidate-piece scores, assigning a fixed low score to candidates unavailable in the returned log-probabilities. For scores $s_{ti}$, the scaled distribution is  
\begin{equation}  
    p^{(T)}_{ti}  
    =  
    \frac{\exp(s_{ti}/T)}  
    {\sum_{k=1}^{24}\exp(s_{tk}/T)}.  
\end{equation}  
The temperature is fitted by minimizing negative log-likelihood on an independent development set. For a fixed candidate set,  
temperature scaling preserves the within-turn ranking of pieces, but may change the ranking of confidence values across turns.  
  
  

\FloatBarrier
\paragraph{Separately elicited beliefs.}  
\begin{figure}[htbp]
\centering
\footnotesize
\begin{minipage}{\linewidth}
\hrule height 0.8pt \vspace{4pt}
\par\vspace{2pt}
\begin{verbatim}
system: You are evaluating which puzzle piece the Helper is most likely
        referring to. There are 24 pieces (#0 through #23). Output a JSON
        object mapping each piece number (string key) to its probability;
        all 24 values must sum to 1.0. Output ONLY the JSON, no markdown.
        Example (if #19 is most likely): {"0":0.01, ..., "19":0.75, ...}
user:   <same flattened context as the action call>
        Now output ONLY the JSON probability distribution over the 24
        pieces (keys "0"-"23", values summing to 1.0). No explanation.
\end{verbatim}
\vspace{2pt}\hrule height 0.8pt
\end{minipage}
\caption{A \emph{separate} \emph{belief-elicitation} call swaps in
a clean system prompt and appends a final JSON instruction, returning a
distribution over the 24 pieces; it is never used to choose the action, letting us
compare represented vs.\ expressed uncertainty.}
\label{fig:belief_elicitation_prompt}
\end{figure}
In a separate model call (See \cref{fig:belief_elicitation_prompt}.), we provide the same pre-action context and request a probability distribution over all 24 candidate pieces. This call does not ask the model to take an action or provide a chain-of-thought explanation. Separating belief elicitation from action generation allows us to compare reported uncertainty with committed behaviour, including cases in which the selected piece differs from the elicited distribution's highest-probability candidate. 

\FloatBarrier
\paragraph{Uncertainty estimation.}  
\label{uncertainty_shannon}
We summarize distributional uncertainty using Shannon entropy:  
\begin{equation}  
    \label{eq:entropy}  
    H_{\mathrm{ref}}(p_t)  
    =  
    -\sum_i p_{ti}\ln p_{ti},  
\end{equation}  
with $p_{ti}$ the probability of piece $i$ in turn $t$. Entropy is measured in nats. Lower values indicate a distribution concentrated on fewer candidates, higher values indicate greater dispersion. Because the signals above differ in construction and candidate coverage, their entropy values are  
not treated as interchangeable scales.

\FloatBarrier
\paragraph{Evaluation metrics.}  
\label{sec:eval_metrics}
For each evaluated placement $t$, let $z_t$ indicate correctness and let $c_t=p_{t,\widehat{y}_t}$ denote the probability assigned to the piece actually selected by the Worker (raw, calibrated or elicited). For human judgments, we use correctness-likelihood ratings rescaled to $[0,1]$, $q_{jt}=r_{jt}/100$, rather than apparent-confidence ratings.  
  
We measure discrimination using AUROC, the probability that a correct selection receives higher confidence than an incorrect selection, with ties receiving half credit~\cite{fawcett2006introduction}:  
\begin{equation}  
    \operatorname{AUROC}  
    =  
    P(c^{+}>c^{-})  
    + \frac{1}{2}P(c^{+}=c^{-}),  
\end{equation}  
where $c^{+}$ and $c^{-}$ denote confidence scores for correct and incorrect selections, respectively. Human AUROC is computed per participant using $q_{jt}$. For entropy-based error detection, we instead use entropy as the score and incorrect selections as the positive class.  
  
The binary Brier score measures probabilistic accuracy ~\cite{glenn1950verification}:  
\begin{equation}  
    \operatorname{BS}  
    =  
    \frac{1}{n}\sum_{t=1}^{n}(c_t-z_t)^2.  
\end{equation}  
Lower values indicate better probabilistic predictions---the Brier score reflects more than calibration alone.  
  
Expected calibration error (ECE) measures the weighted average absolute difference between empirical accuracy  
and mean confidence within confidence bins~\cite{naeini2015obtaining}:  
\begin{equation}  
    \operatorname{ECE}  
    =  
    \sum_{m=1}^{M}  
    \frac{|B_m|}{n}  
    \left|  
        \operatorname{acc}(B_m)  
        -  
        \operatorname{conf}(B_m)  
    \right|,  
\end{equation}  
where $\operatorname{acc}(B_m)$ and $\operatorname{conf}(B_m)$ are the mean correctness and confidence in bin $B_m$. Empty bins contribute zero.  
Lower ECE indicates closer agreement at the chosen binning resolution. Human ECE is computed over pooled  participant--turn ratings, substituting $q_{jt}$ for $c_t$.   

\FloatBarrier
\subsection{Models and Prompt Format}  
\label{sec:models}  
\begin{figure}[t]
\centering
\footnotesize
\begin{minipage}{\linewidth}
\hrule height 0.8pt \vspace{4pt}
\par\vspace{2pt}
\begin{verbatim}
system: <Worker system prompt from the benchmark dialogue>
        role: sees the 24 staged pieces, CANNOT see the target, must rely
        on the Helper. ACTION FORMAT: "PLACE piece #N at (x, y)" (or
        "MOVE piece from (x1,y1) to (x2,y2)"), always with a MESSAGE: "...".
user:   <one flattened message; only the CURRENT turn carries board+pieces>
===================== Conversation History =====================
[turn 1] HELPER: <instruction text only>      # "--- Pieces ---" stripped
[turn 2] WORKER: <reply text only>
   ...
===================== Current Turn (Your instructions) =====================
<current Helper instruction>  +  <board & 24 pieces as IMAGE (or TEXT)>
\end{verbatim}
\vspace{2pt}\hrule height 0.8pt
\end{minipage}
\caption{\textbf{Off-policy single-prompt format.} Each recorded turn is scored
with two independent calls that share the same flattened context. The dialogue up
to the current turn is collapsed into one user message: earlier turns as a
transcript (\emph{Conversation History}, instruction text only), and the current
Helper instruction (\emph{Current Turn}) carrying the board and 24 pieces as an
image or as text. The \emph{action} call selects and places a piece
(temperature $0$; for \textsc{GPT-4.1} we read the top-20 log-probs at the
piece-number token).}
\label{fig:single_prompt}
\end{figure}
  
We evaluate \textsc{GPT-4.1 (2025-04-14)}, \textsc{GPT-5 (2025-08-07)}, and \textsc{GPT-5.5 (2026-04-24)} through Azure OpenAI. Action-token log-probabilities are available for GPT-4.1 but not for GPT-5 or GPT-5.5 in our evaluation setup. Cross-model comparisons therefore use the separately elicited  belief distribution. For GPT-4.1 we use temperature $0$, for this parameter is not available and we set the thinking effort to default. 
  
To compare models without collecting new live interactions, we replay individual Helper turns using a single prompt (See \cref{fig:single_prompt}.). The preceding dialogue is included under \emph{Conversation History}, followed by the Helper's instruction under \emph{Current Turn}. The Worker's board and candidate pieces are supplied either as images or as textual descriptions.  
  
This is an off-policy evaluation in which each model responds to a recorded history that may have been generated by a different model. Its response is not propagated into subsequent turns. The design isolates piece-level decisions under matched contexts, but does not capture how alternative actions or clarification requests would change the subsequent conversation. For validation, single-prompt GPT-4.1 replay agrees with the original multi-turn behaviour on whether to act in $95\%$ of evaluated cases and on the selected piece in $86\%$ of cases.

\FloatBarrier
\subsection{Statistical Analysis Details}  
\label{sec:stats}  

\FloatBarrier
\paragraph{Offline model comparisons.}  
For matched image--text evaluations, we compare placement correctness using exact McNemar tests on turns with placements under both modalities, and belief entropy using Wilcoxon signed-rank tests. We report coverage separately to distinguish accuracy differences from differences in the willingness to act.  
  
For ordered instruction-specificity and board-confusability conditions, we estimate linear trends separately by model and modality. Accuracy and coverage are analysed using linear-probability models, and entropy using ordinary least squares. Standard errors are clustered by source dialogue  
to account for repeated observations from the same interaction.  
  
We evaluate natural-turn hedge triggering by using elicited belief entropy to predict incorrect placements scored against  
ground truth piece. We report AUROC and, for a specified threshold, error recall, false-positive rate, and precision.  
These analyses distinguish whether uncertainty tracks controlled  changes in task difficulty from whether it identifies individual  
errors in unmodified interactions.  

\FloatBarrier
\paragraph{Human-study comparisons.}  
The participant is the unit of inference for condition comparisons. We aggregate repeated turn-level responses within participants and compare participant summaries using permutation tests. We report effect sizes and percentile-bootstrap confidence intervals using $10{,}000$ resamples unless otherwise stated.  
  
We group comparisons among Generic, Described, Hedged, and  Visible board into a primary family, and comparisons of Self-hedged with the other conditions into a separate exploratory family. Holm correction controls family-wise error within the specified comparison families. Questionnaire analyses and allocation checks are reported separately.  
  
  
For the targeted hedging analysis, we compare judgments on flagged and un-flagged turns, further separated by ground-truth  correctness. We additionally use an exact dialogue-level sign-flip test for the specified mechanism contrast to assess whether the effect is consistent across source dialogues.  

See also \cref{tab:statistical_tests} for a summary of all used statistical tests.

\FloatBarrier
\subsection{Additional Calibration Results}
\begin{table}[htbp]
\centering\small
\caption{Calibration and discrimination with 95\% bootstrap CIs (20{,}000 resamples) and significance, for the same $n{=}554$ committed GPT-4.1 placements (image, on-policy L0; 296 correct, 258 wrong). Overconfidence $=$ mean confidence $-$ accuracy (calibration-in-the-large); the AUROC $p$-value is a two-sided Mann–Whitney test against chance ($0.5$).}
\label{tab:s1_calibration_full}
\begin{tabular}{lccccc}
\toprule
Signal & Mean conf. & Overconfidence & AUROC ($p$ vs.\ .5) & ECE & Brier \\
\midrule
Raw log-prob                & $0.973$ & $+0.438\,[0.40,0.48]$ & $0.644\,[0.59,0.68]$ $(p{<}10^{-7})$ & $0.442\,[0.40,0.48]$ & $0.439\,[0.40,0.48]$ \\
Calibrated ($T^\star{=}16.5$)  & $0.531$ & $-0.003\,[\text{-}0.05,0.04]$ & $0.623\,[0.58,0.67]$ $(p{<}10^{-6})$ & $0.157\,[0.13,0.21]$ & $0.271\,[0.25,0.30]$ \\
Elicited belief             & $0.547$ & $+0.013\,[\text{-}0.03,0.05]$ & $\mathbf{0.648}\,[0.61,0.70]$ $(p{<}10^{-9})$ & $\mathbf{0.152}\,[0.13,0.20]$ & $\mathbf{0.253}\,[0.23,0.28]$ \\
\bottomrule
\end{tabular}
\end{table}

\begin{table}[htbp]
\centering\small
\caption{Pairwise AUROC comparisons (paired bootstrap, $20{,}000$ resamples, $n{=}554$); positive $\Delta$AUROC favours the first signal.}
\label{tab:s1_auroc_pairwise}
\begin{tabular}{lcc}
\toprule
Comparison & $\Delta$AUROC [95\% CI] & $p$ \\
\midrule
Elicited belief $-$ Calibrated log-prob & $+0.025\,[-0.02,+0.08]$ & $.26$ \\
Elicited belief $-$ Raw log-prob        & $+0.003\,[-0.03,+0.07]$ & $.41$ \\
Raw log-prob $-$ Calibrated log-prob    & $+0.021\,[-0.01,+0.03]$ & $.46$ \\
\bottomrule
\end{tabular}
\end{table}

\FloatBarrier
\subsection{Additional Robustness Results}
\FloatBarrier
\paragraph{Modality}
\begin{figure}[htbp] 
    \centering  
    \includegraphics[width=0.8\textwidth]  
        {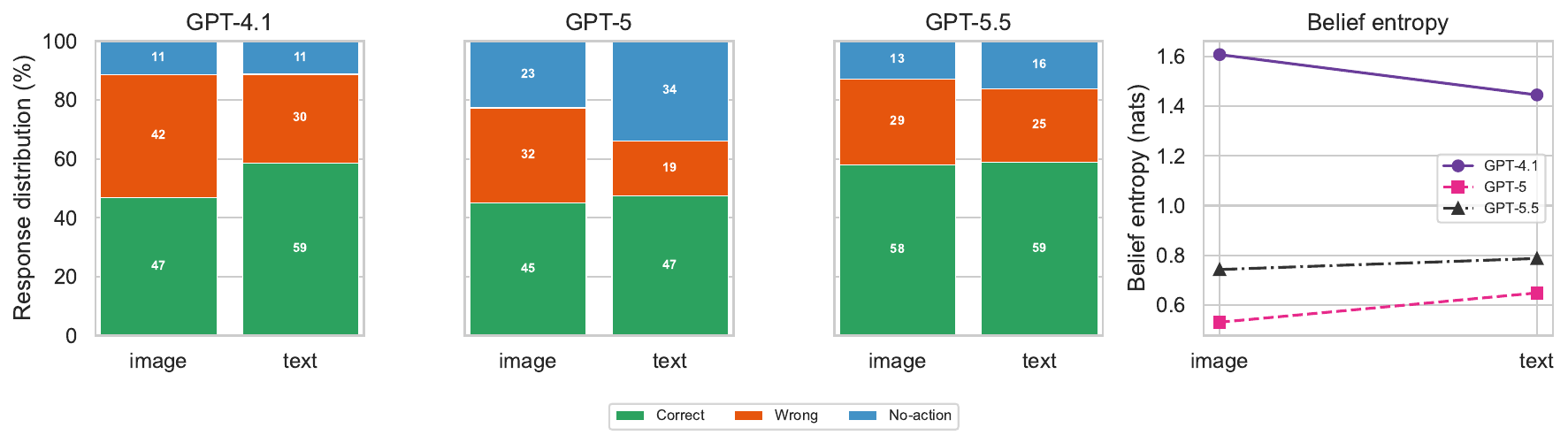}  
    \caption{Context modality (image vs.\ text) results per model. (left) stacked
outcome composition of Worker responses (correct piece / wrong piece / no-action); (right) mean belief entropies. Text improves accuracy for GPT-4.1 and GPT-5 and leaves GPT-5.5 unchanged, while belief entropy moves little and inconsistently.}  
    \label{fig:s2b_modality}  
\end{figure} 

\begin{figure}[htbp]
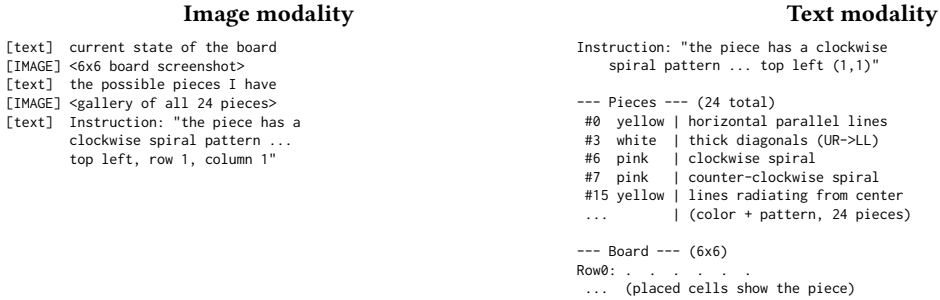

\centering
\begin{minipage}[t]{0.46\linewidth}
\centering\textbf{Image modality}\par\vspace{2pt}
\scriptsize
\begin{verbatim}
[text]  current state of the board
[IMAGE] <6x6 board screenshot>
[text]  the possible pieces I have
[IMAGE] <gallery of all 24 pieces>
[text]  Instruction: "the piece has a
        clockwise spiral pattern ...
        top left, row 1, column 1"
\end{verbatim}
\end{minipage}\hfill
\begin{minipage}[t]{0.50\linewidth}
\centering\textbf{Text modality}\par\vspace{2pt}
\scriptsize
\begin{verbatim}
Instruction: "the piece has a clockwise
    spiral pattern ... top left (1,1)"

--- Pieces --- (24 total)
 #0  yellow | horizontal parallel lines
 #3  white  | thick diagonals (UR->LL)
 #6  pink   | clockwise spiral
 #7  pink   | counter-clockwise spiral
 #15 yellow | lines radiating from center
 ...        | (color + pattern, 24 pieces)

--- Board --- (6x6)
Row0: .  .  .  .  .  .
 ...  (placed cells show the piece)
\end{verbatim}
\end{minipage}
\caption{\textbf{The same turn in two context modalities.} Identical instruction,
board, and 24 pieces, rendered either \textbf{(left)} as two images---a board
screenshot and a gallery of all pieces---that the Worker must read visually, or
\textbf{(right)} as text, with each piece a \texttt{color | pattern} line and the board
an ASCII grid. Verbal descriptions preserve distinctions that images blur: the target
\texttt{\#6 clockwise spiral} and its near-twin \texttt{\#7 counter-clockwise spiral}
are a confusable pair by sight but separable in words. This is the manipulation behind
\cref{fig:s2b_modality}. (Piece list abridged to $5$ of $24$ for space.)}
\label{fig:modality_example}
\end{figure}

\begin{table}[htbp]
\centering
\caption{Context modality (image vs.\ text), paired within model on matched dialogue turns.
On-action accuracy is compared with an exact McNemar test on turns where the model committed a
placement in both modalities; belief entropy with a Wilcoxon signed-rank test (effect size =
matched-pairs rank-biserial $r$). No-action rate is reported descriptively; whether and how much
models ask is analysed in \cref{sec:study3}.}
\label{tab:s2b_modality}
\begin{tabular}{lcccccc}
\toprule
& \multicolumn{2}{c}{On-action acc.\ (\%)} & No-action (\%) & \multicolumn{3}{c}{Belief entropy (nats)}\\
\cmidrule(lr){2-3}\cmidrule(lr){5-7}
Model & img $\to$ txt & McNemar $p$ & img $\to$ txt & img $\to$ txt & $r$ & Wilcoxon $p$\\
\midrule
GPT-4.1 & $52.9 \to 66.0$ & $1.1\times10^{-7}$ & $11.4 \to 11.2$ & $1.74 \to 1.56$ & $+0.36$ & $2.7\times10^{-14}$\\
GPT-5   & $58.4 \to 71.7$ & $3.5\times10^{-9}$ & $22.7 \to 33.9$ & $0.78 \to 0.84$ & $-0.23$ & $4.2\times10^{-7}$\\
GPT-5.5 & $66.7 \to 70.1$ & $0.16$            & $12.9 \to 16.1$ & $0.95 \to 0.95$ & $+0.02$ & $0.62$\\
\bottomrule
\end{tabular}
\end{table}

\FloatBarrier
\paragraph{Instructions}
\begin{table}[htbp]
\centering
\small
\caption{Instruction-specificity levels. The synthetic gradient replaces the Helper's
message with a deterministic template that strips referential cues step by step, from
piece id\,$+$\,full description\,$+$\,position down to a generic pointer. Specificity is
ordinal across \textbf{L1}$\rightarrow$\textbf{L4} (most $\rightarrow$ least informative);
\textbf{L0} is the real human message, placed mid-axis (between L2 and L3) as the baseline. A further misleading level
(L5), which describes a confusable piece, is reported in the appendix. Running example:
target $=$ the pink clockwise-spiral piece (\#6).}
\label{tab:levels}
\begin{tabular}{@{}l p{3.1cm} p{7.1cm}@{}}
\toprule
\textbf{Level} & \textbf{Cues provided} & \textbf{Example instruction} \\
\midrule
\textbf{L0}\, Original       & original human message                 & \emph{``the piece has a clockwise spiral pattern \ldots\ top-left, row~1, column~1''} \\
\textbf{L1}\, Over-specified & piece \#, colour, pattern, position    & ``Place piece \#6 at (2,\,3). It's the pink piece with clockwise spiral emanating from center.'' \\
\textbf{L2}\, Feature-rich   & colour, pattern, position              & ``Find the pink piece that has clockwise spiral emanating from center. Place it at (2,\,3).'' \\
\textbf{L3}\, Single feature & colour only                            & ``Place the pink piece.'' \\
\textbf{L4}\, Vague          & no distinguishing feature              & ``Place the next piece.'' \\
\bottomrule
\end{tabular}
\end{table}

\begin{table}[htbp]
\centering
\caption{Instruction-specificity templates (pseudo-code); each level overwrites the current Helper
instruction. Placeholders \texttt{\{id\}}, \texttt{\{color\}}, \texttt{\{pattern\}}, \texttt{\{pos\}}
are filled from the target piece (e.g.\ piece \#6 $=$ ``pink'', ``clockwise spiral emanating from
center''); \texttt{\{pos\}} is a fixed placeholder because only referent identity is varied. L0 is the
untouched human instruction (mid-axis baseline); a further ``misleading'' template describing a
confusable competitor exists in code but is not part of the reported axis.}
\label{tab:instr_templates}
\begin{minipage}{0.92\linewidth}
\footnotesize
\begin{verbatim}
GenerateInstruction(piece, level):
    id, color, pattern <- attributes(piece)   # "pink", "clockwise spiral ..."
    pos <- "(2,3)"                            # fixed: only referent identity varies

    L1  over-specified :  "Place piece #{id} at {pos}. "
                          "It's the {color} piece with {pattern}."
    L2  feature-rich   :  "Find the {color} piece that has {pattern}. "
                          "Place it at {pos}."
    L3  single feature :  "Place the {color} piece."
    L4  vague          :  "Place the next piece."
    L0  baseline       :  <original human instruction, unchanged>
\end{verbatim}
\end{minipage}
\end{table}

For the target piece \#6 (pink; ``clockwise spiral emanating from center'') these instantiate to:
L1 ``Place piece \#6 at (2,\,3). It's the pink piece with clockwise spiral emanating from center.'';
L2 ``Find the pink piece that has clockwise spiral emanating from center. Place it at (2,\,3).'';
L3 ``Place the pink piece.''; and L4 ``Place the next piece.'

\begin{figure}[htbp]  
    \centering  
    \includegraphics[width=1\textwidth]  
        {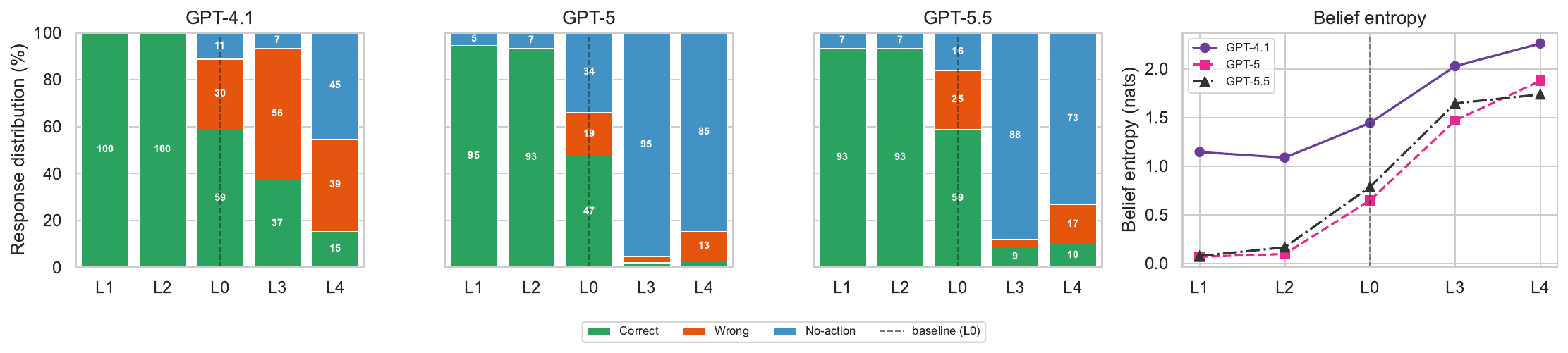}  
    \caption{Instruction specificity (text modality) results per model. (left) stacked outcome composition of Worker response (correct piece / wrong piece / no-action) at each specificity level from overspecified (L1) to vague (L4), with the original utterance (L0) placed between L2 and L3 as the baseline (dashed); the right panel shows mean belief entropy. Accuracy falls and belief entropy rises with vagueness for all models.}
    \label{fig:s2a_levels_text}  
\end{figure} 

\begin{table}[htbp]
\centering\small
\caption{Instruction specificity: linear trend in each outcome per step along the ordered
axis L1\,$\to$\,L2\,$\to$\,L0\,$\to$\,L3\,$\to$\,L4 (increasing vagueness; the original
utterance L0 sits mid-axis). Per-model regressions with dialogue-clustered standard errors
(25 dialogues); accuracy and coverage are linear-probability models (percentage points per
step), belief entropy is OLS (nats per step). All slopes $p<0.001$.}
\label{tab:s2a_trend}
\begin{tabular}{@{}llccc@{}}
\toprule
Model & Modality & Accuracy (pp/step) & Coverage (pp/step) & Entropy (nats/step)\\
\midrule
GPT-4.1 & image & $-21.2\,[-22.9, -19.6]$ & $-9.3\,[-11.0, -7.5]$ & $+0.313\,[+0.275, +0.351]$\\
        & text  & $-20.6\,[-22.4, -18.8]$ & $-9.7\,[-12.2, -7.3]$ & $+0.317\,[+0.297, +0.338]$\\
GPT-5   & image & $-20.2\,[-23.6, -16.7]$ & $-24.3\,[-25.5, -23.2]$ & $+0.510\,[+0.471, +0.550]$\\
        & text  & $-18.4\,[-22.6, -14.3]$ & $-24.7\,[-26.3, -23.1]$ & $+0.498\,[+0.459, +0.538]$\\
GPT-5.5 & image & $-15.5\,[-19.1, -12.0]$ & $-23.2\,[-25.2, -21.2]$ & $+0.469\,[+0.434, +0.503]$\\
        & text  & $-16.4\,[-20.5, -12.3]$ & $-21.5\,[-23.6, -19.4]$ & $+0.480\,[+0.446, +0.515]$\\
\bottomrule
\end{tabular}
\end{table}

\FloatBarrier
\paragraph{Confusability}
\begin{figure}[htbp]
    \centering
    \includegraphics[width=\linewidth]{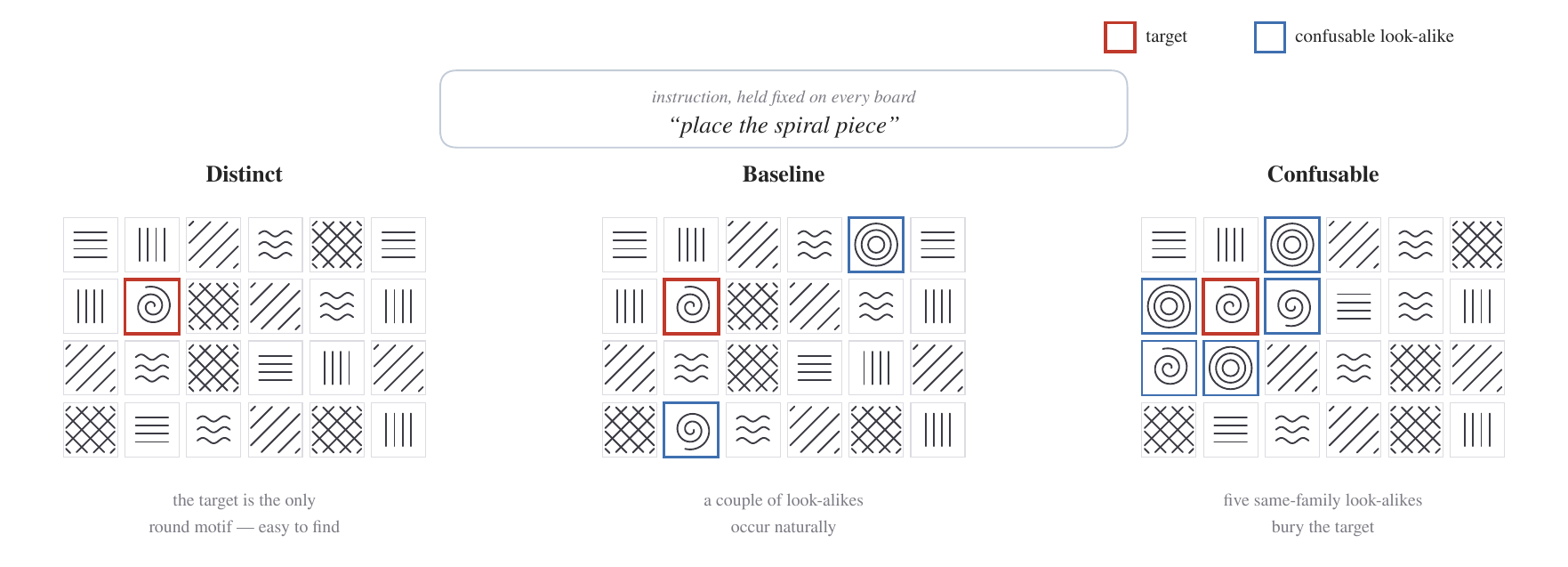}
    \caption{\textbf{Board (environmental) confusability.} The \emph{same} fixed
    referring instruction (``place the spiral piece'') is shown against three boards
    along a confusability axis; the instruction is held constant so that only the
    competing referents change. \textbf{Distinct (B1):} the target is the only piece of
    its shape family and thus the sole plausible referent. \textbf{Baseline (B0):} the
    real board seen in the dialogue, where a couple of incidental look-alikes occur
    naturally. \textbf{Confusable (B2):} five same-family, same-thickness look-alikes are
    inserted, burying the target among near-identical distractors. The red ring marks the
    intended target; the blue ring marks a confusable same-family look-alike. Moving from
    distinct to confusable makes it increasingly likely that the Worker commits to a
    look-alike---the environmental (referent-side) axis of referential uncertainty,
    complementary to instruction vagueness (\cref{fig:s2a_levels}).}
    \label{fig:board_confusability}
\end{figure}

\begin{figure}[t]  
    \centering  
    \includegraphics[width=1\textwidth]  
        {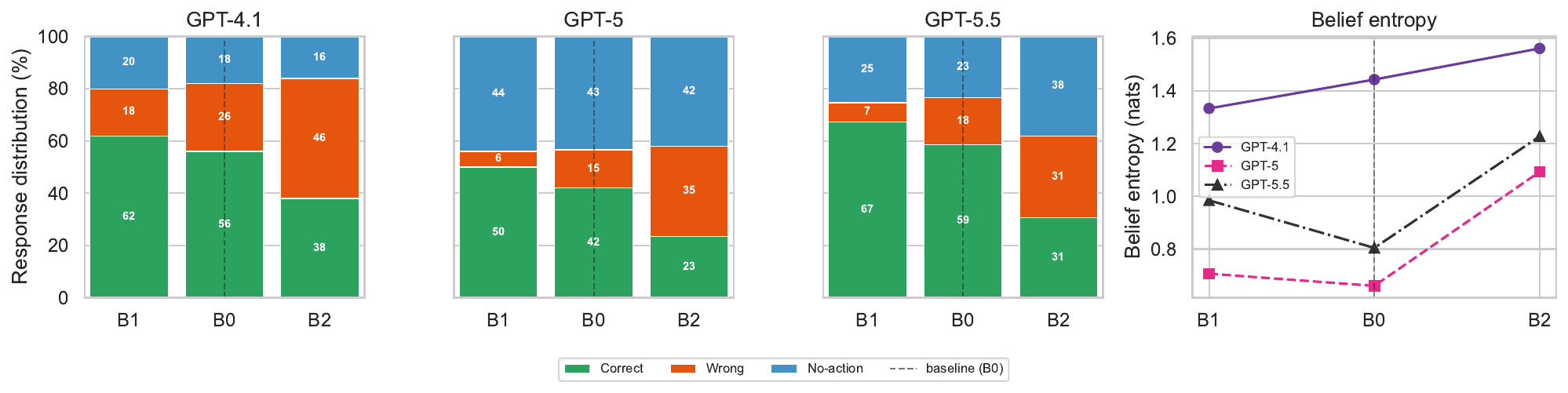}  
    \caption{Board confusability (text modality) results per model. (left) stacked outcome
composition of Worker responses (correct piece / wrong piece / no-action) and (right) mean belief entropy on a distinct board (B1), the original board (B0), and a confusable board (B2), in order of increasing confusability. Accuracy falls sharply while no-action and belief entropy barely move.}  
    \label{fig:s2c_boards_text}  
\end{figure}

\begin{table}[htbp]\centering\small
\caption{Board confusability: linear trend in each outcome per step along the axis
B1 (distinct)\,$\to$\,B0 (original)\,$\to$\,B2 (confusable); per-model regressions with
dialogue-clustered SEs (23 dialogues). Accuracy/coverage in percentage points per step,
belief entropy in nats per step. All accuracy and entropy trends are significant
($p\le.002$); coverage is non-significant except GPT-5.5 text (\textbf{bold}, $p<.001$).}
\label{tab:s2c_trend}
\begin{tabular}{@{}llccc@{}}
\toprule
Model & Modality & Accuracy (pp/step) & Coverage (pp/step) & Entropy (nats/step)\\
\midrule
GPT-4.1 & image & $-11.7\,[-15.3,-8.1]$  & $+0.3\,[-0.8,+1.5]$   & $+0.052\,[+0.019,+0.085]$\\
        & text  & $-16.2\,[-21.4,-10.9]$ & $+2.0\,[-0.3,+4.3]$   & $+0.114\,[+0.073,+0.154]$\\
GPT-5   & image & $-15.9\,[-21.8,-10.0]$ & $+1.3\,[-2.1,+4.8]$   & $+0.063\,[+0.037,+0.090]$\\
        & text  & $-24.6\,[-31.5,-17.7]$ & $+1.0\,[-4.3,+6.3]$   & $+0.194\,[+0.136,+0.251]$\\
GPT-5.5 & image & $-15.8\,[-20.8,-10.9]$ & $+1.7\,[-0.6,+3.9]$   & $+0.063\,[+0.028,+0.097]$\\
        & text  & $-20.1\,[-25.0,-15.2]$ & $\mathbf{-6.3\,[-10.0,-2.7]}$ & $+0.122\,[+0.074,+0.169]$\\
\bottomrule
\end{tabular}
\end{table}

\FloatBarrier
\subsection{Additional Externalizing Uncertainty Results}
\begin{table}[htbp]\centering\small
\caption{Judged clarification rate per model and modality. Each no-action is labeled by an LLM judge
(GPT-5.5) as an explicit clarification (names the confusing piece), a generic hedge, or a non-placement
turn; the rate is (explicit\,+\,hedge) over placement-instruction turns (all turns not labeled NOT\_PLACEMENT).}
\label{tab:s3a_clarify}
\begin{tabular}{@{}llcccc@{}}
\toprule
Model & Modality & $N$ & Explicit (\%) & Hedge (\%) & Clarification (\%)\\
\midrule
GPT-4.1 & image & 546 & 2.93  & 0.55 & 3.48\\
        & text  & 561 & 1.25  & 0.36 & 1.60\\
GPT-5   & image & 556 & 15.83 & 0.90 & 16.73\\
        & text  & 562 & 23.13 & 0.71 & 23.84\\
GPT-5.5 & image & 549 & 5.65  & 0.55 & 6.19\\
        & text  & 555 & 8.47  & 0.36 & 8.83\\
\bottomrule
\end{tabular}
\end{table}

\begin{table}[htbp]\centering\small
\caption{Is clarification targeted at error-prone turns? Clarification rate on turns where the recorded
AI Worker (reproduced by GPT-4.1) placed the \emph{wrong} piece (needed) vs.\ the \emph{correct} piece
(not needed); one-sided Fisher exact test ($H_1$: higher when needed).}
\label{tab:s3a_targeting}
\begin{tabular}{@{}llccc@{}}
\toprule
Model & Modality & Needed (\%, $n$) & Not needed (\%, $n$) & Fisher $p$\\
\midrule
GPT-4.1 & image & 1.98 (101)  & 1.74 (115)  & 0.64\\
        & text  & 1.94 (103)  & 0.85 (117)  & 0.45\\
GPT-5   & image & 22.33 (103) & 11.11 (117) & \textbf{0.020}\\
        & text  & 29.41 (102) & 26.50 (117) & 0.37\\
GPT-5.5 & image & 2.94 (102)  & 3.45 (116)  & 0.72\\
        & text  & 10.58 (104) & 5.22 (115)  & 0.11\\
\bottomrule
\end{tabular}
\end{table}

\begin{table}[htbp]\centering\small
\caption{Externalizing uncertainty on natural turns. \textbf{Q1 (coupling):} point-biserial correlation between
questioning (no-action) and elicited belief entropy on graded beliefs, with mean entropy on ask vs.\ place
turns (all $p<10^{-4}$). \textbf{Q2 (hedge trigger):} on acted turns scored against \texttt{ground\_truth\_piece},
AUROC of belief entropy predicting a wrong placement, and the recall / false-hedge rate (FPR) / precision of the
rule ``hedge if entropy $\geq$ median'' (base error rate last).}
\label{tab:s3_trigger}
\begin{tabular}{@{}llccc c ccccc@{}}
\toprule
& & \multicolumn{3}{c}{Q1: coupling} & & \multicolumn{5}{c}{Q2: hedge trigger (vs.\ GT)}\\
\cmidrule(lr){3-5}\cmidrule(lr){7-11}
Model & Mod. & $r_{\text{ask},H}$ & $H_{\text{ask}}$ & $H_{\text{place}}$ & & AUROC & recall & FPR & prec. & base err.\\
\midrule
GPT-4.1 & image & 0.18 & 1.87 & 1.58 & & 0.548 & 0.57 & 0.46 & 0.53 & 47.1\%\\
        & text  & 0.31 & 1.82 & 1.43 & & 0.602 & 0.62 & 0.48 & 0.40 & 34.0\%\\
GPT-5   & image & 0.35 & 0.99 & 0.49 & & 0.603 & 0.57 & 0.46 & 0.47 & 41.6\%\\
        & text  & 0.39 & 1.04 & 0.54 & & 0.666 & 0.67 & 0.44 & 0.38 & 28.3\%\\
GPT-5.5 & image & 0.31 & 1.16 & 0.72 & & 0.659 & 0.68 & 0.41 & 0.45 & 33.3\%\\
        & text  & 0.31 & 1.19 & 0.75 & & 0.611 & 0.63 & 0.45 & 0.38 & 29.9\%\\
\bottomrule
\end{tabular}
\end{table}

\FloatBarrier
\subsection{Grading Interface}
\begin{figure}[htbp]\centering
\begin{tikzpicture}
  \node[anchor=south west,inner sep=0] (img)
     {\includegraphics[width=0.82\linewidth]{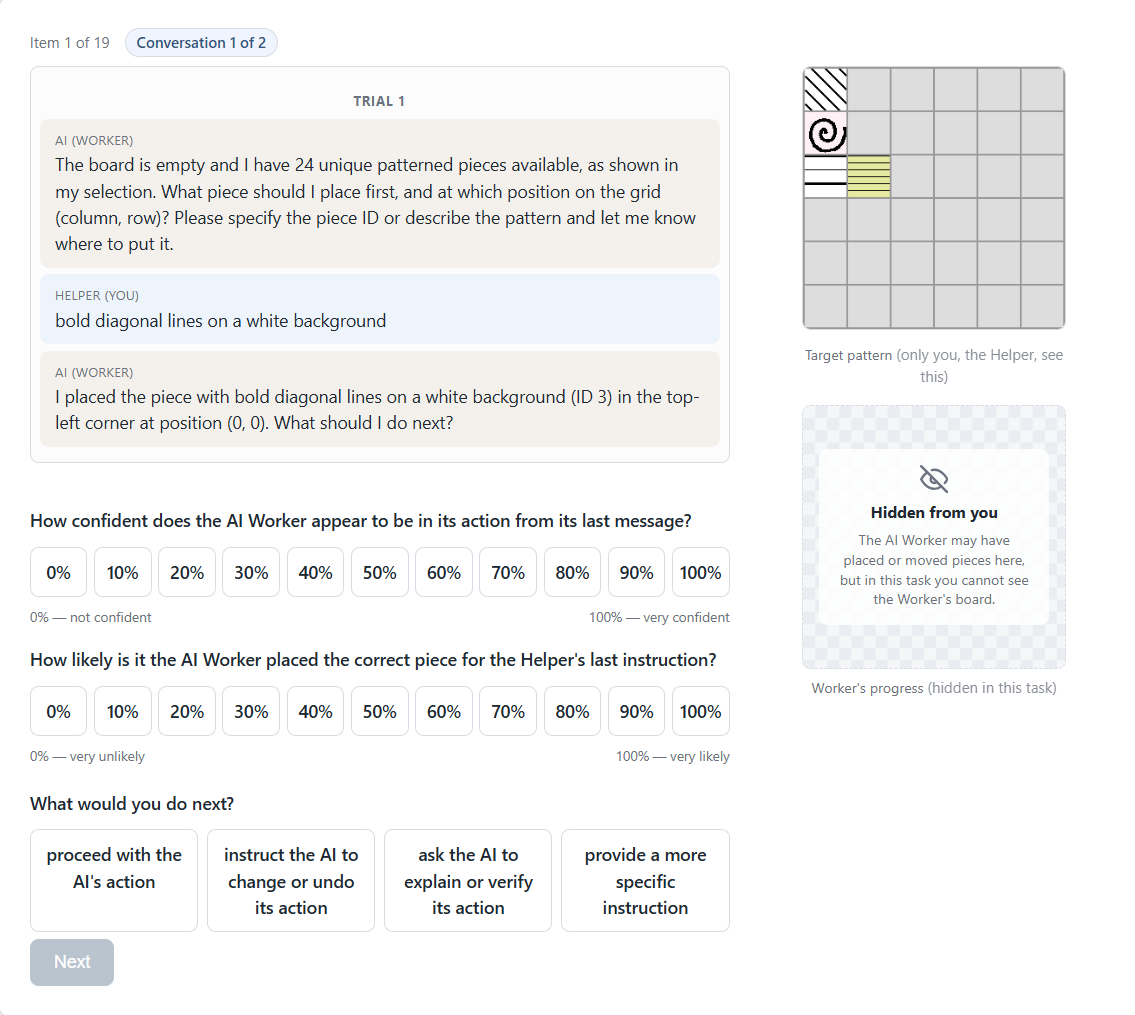}};
  \begin{scope}[x={(img.south east)},y={(img.north west)},
                font=\sffamily\bfseries\small]
    \node[circle,draw,fill=white,inner sep=1pt] at (0.03,0.86) {a}; 
    \node[circle,draw,fill=white,inner sep=1pt] at (0.03,0.46) {b}; 
    \node[circle,draw,fill=white,inner sep=1pt] at (0.03,0.33) {c}; 
    \node[circle,draw,fill=white,inner sep=1pt] at (0.03,0.16) {d}; 
    \node[circle,draw,fill=white,inner sep=1pt] at (0.72,0.86) {e}; 
  \end{scope}
\end{tikzpicture}
\caption{\textbf{The grading interface (one trial, \emph{Generic} condition).} In this between-subjects study, each participant reviews the flattened interaction up to the current turn and judges the AI Worker's placement. \textbf{(a)}~The dialogue---the Worker's opening status, the Helper's instruction (here \emph{``bold diagonal lines on a white background''}), and the Worker's placement message. For that message the participant reports two $0$--$100$ judgements, \textbf{(b)}~\emph{apparent confidence} (how confident the Worker sounds) and \textbf{(c)}~\emph{correctness likelihood} (how likely the intended piece was placed), and \textbf{(d)}~chooses a single \emph{action}: proceed, ask the Worker to change/undo, ask it to explain/verify, or give a more specific instruction. \textbf{(e)}~As the Helper, the participant sees the target pattern but \emph{not} the Worker's board; the board is revealed only in the \emph{Visible-board} condition. Across conditions we vary only the Worker's message and board visibility (\cref{tab:conditions_primary}), holding the task turns fixed. The available pieces are not shown in any condition as they are also not available to see for the Helper.}
\label{fig:grader_ui}
\end{figure}

\FloatBarrier
\subsection{Human Perception Study Results Details}
\begin{table}[htbp]
\centering
\small
\caption{Human study per-condition results---full descriptive breakdown (participant means; $N=210$,
${\sim}42$/condition). Perception ratings are on a $0$--$100$ scale; the action mix gives the
percentage of substantive turns assigned to each of the four mutually exclusive responses (columns
sum to $100\%$ up to rounding); error sensitivity gives acceptance rates conditioned on ground-truth
placement correctness. \emph{Rating discrimination} $=$ mean(correctness $\mid$ correct) $-$
mean(correctness $\mid$ wrong); \emph{action discrimination} $=$ $P(\text{accept}\mid\text{correct})
- P(\text{accept}\mid\text{wrong})$. Self-hedged is the secondary, deployable condition.}
\label{tab:s4_descriptives_full}
\begin{tabular}{lccccc}
\toprule
& Generic & Described & Hedged & Visible board  & Self-\\hedged \\
\midrule
\multicolumn{6}{l}{\emph{Perception (observer ratings, 0--100)}}\\
\quad Correctness-likelihood                 & 77.6 & 67.3 & 56.1 & 53.7 & 61.5 \\
\quad Apparent confidence                    & 85.5 & 85.7 & 58.5 & 83.3 & 56.6 \\
\quad Rating discrimination (corr.$-$wrong)  & $-0.1$ & 16.8 & 27.7 & 26.6 & 8.8 \\
\midrule
\multicolumn{6}{l}{\emph{Action mix (\% of substantive turns)}}\\
\quad Proceed / accept                       & 76.7 & 63.8 & 50.6 & 41.5 & 54.4 \\
\quad Change / undo                          & 6.5  & 17.5 & 20.1 & 26.3 & 16.9 \\
\quad Ask to explain / verify                & 7.1  & 8.0  & 12.3 & 10.1 & 10.2 \\
\quad More-specific instruction              & 9.7  & 10.8 & 17.0 & 22.0 & 18.5 \\
\midrule
\multicolumn{6}{l}{\emph{Error sensitivity (\% accepted)}}\\
\quad Accept correct moves                   & 79.0 & 78.6 & 73.7 & 64.8 & 60.1 \\
\quad Accept wrong moves                     & 78.3 & 57.1 & 36.4 & 30.1 & 50.2 \\
\quad Action discrimination (pp)             & 0.7  & 21.5 & 37.3 & 34.8 & 10.0 \\
\midrule
$N$ participants                             & 41 & 42 & 42 & 43 & 42 \\
\bottomrule
\end{tabular}
\end{table}

\begin{table}[htbp]
\centering
\small
\caption{Human post-study questionnaire (participant means; $N=210$, all responded).
Prior-AI items are pre-existing attitudes and act as a randomisation check (balanced across
conditions); pair items are the condition-dependent subjective experience. ``Helper repeated
explanations'' is scored higher $=$ more repetition; the pair composite reverse-scores it ($6-x$)
so higher $=$ better. The conditions with the best decision quality (Hedged, Visible board;
\cref{tab:s4_results}) yield the \emph{lowest} subjective smoothness, whereas Generic---a
near-perfect rubber stamp---feels the most fluent.}
\label{tab:s4_survey}
\begin{tabular}{llccccc}
\toprule
& Scale & Generic & Described & Hedged & Visible board & Self-hedged \\
\midrule
\multicolumn{7}{l}{\emph{Prior-AI attitudes (randomisation check)}}\\
\quad Puzzle familiarity           & 1--7 & 3.59 & 3.81 & 2.93 & 3.53 & 3.14 \\
\quad Trust AI models              & 1--5 & 3.58 & 3.67 & 3.60 & 3.47 & 3.43 \\
\quad Understand why AI answered   & 1--5 & 3.88 & 3.93 & 3.60 & 3.74 & 3.71 \\
\quad Detect AI uncertainty        & 1--5 & 3.92 & 3.64 & 3.57 & 3.67 & 3.83 \\
\quad Prior-AI composite           & 1--5 & 3.79 & 3.75 & 3.59 & 3.63 & 3.66 \\
\midrule
\multicolumn{7}{l}{\emph{Interaction experience (pair items)}}\\
\quad Pair on same page            & 1--5 & 3.98 & 3.48 & 3.36 & 2.95 & 3.38 \\
\quad AI understood Helper         & 1--5 & 4.08 & 3.60 & 3.24 & 3.30 & 3.24 \\
\quad Interaction was smooth       & 1--5 & 4.08 & 3.38 & 3.29 & 3.00 & 3.33 \\
\quad Helper repeated explanations & 1--5 & 2.55 & 2.60 & 3.24 & 2.88 & 3.21 \\
\quad Pair composite               & 1--5 & 3.89 & 3.46 & 3.16 & 3.09 & 3.18 \\

\bottomrule
\end{tabular}
\end{table}

\FloatBarrier
\subsection{Statistical Tests Summary}
\begin{table}[htbp]  
\centering  
\small  
\caption{Statistical tests and inferential procedures used in the manuscript.  
Section references indicate the main analyses; implementation details are  
provided in \cref{sec:stats}.}  
\label{tab:statistical_tests}  
\begin{tabular}{@{}p{3.3cm}p{4.8cm}p{4.3cm}@{}}  
\toprule  
\textbf{Test / procedure} & \textbf{Description} & \textbf{Applied} \\  
\midrule  
  
Mann--Whitney $U$ test (two-sided)  
& Test whether confidence discriminates correct from incorrect placements  
(AUROC above chance).  
& \cref{sec:study1}; Appendix,  
\cref{tab:s1_calibration_full}. \\[3pt]  
  
Paired bootstrap  
& Compare AUROC between uncertainty signals; estimate confidence  
intervals and $p$-values.  
& \cref{sec:study1}; Appendix,  
\cref{tab:s1_auroc_pairwise}. \\[3pt]  
  
Exact McNemar test  
& Compare placement correctness between image and text on matched  
turns with placements in both modalities.  
& \cref{sec:s2_modality}; Appendix,  
\cref{tab:s2b_modality}. \\[3pt]  
  
Wilcoxon signed-rank test  
& Compare belief entropy between matched image and text turns.  
& \cref{sec:s2_modality}; Appendix,  
\cref{tab:s2b_modality}. \\[3pt]  
  
Regression trend tests  
(dialogue-clustered SEs)  
& Test ordered trends in accuracy and coverage  
(linear-probability models) and entropy (OLS).  
& \cref{sec:s2_instruction}  
and \cref{sec:s2_clusters}; Appendix,  
\cref{tab:s2a_trend} and~\ref{tab:s2c_trend}. \\[3pt]  
  
Binomial test  
& Test whether belief-argmax switches land within the confusable  
cluster more often than chance.  
& \cref{sec:s2_clusters};  
\cref{tab:2c_conf_switches}. \\[3pt]  
  
Fisher's exact test (one-sided)  
& Test whether clarification is more frequent on recorded incorrect  
than correct placements.  
& \cref{sec:s3_clarify}; Appendix,  
\cref{tab:s3a_targeting}. \\[3pt]  
  
Point-biserial correlation significance test  
& Test the association between no-action/questioning and belief entropy.  
& \cref{sec:s3_trigger}; Appendix,  
\cref{tab:s3_trigger}. \\[3pt]  
  
Participant-level permutation tests  
& Compare condition means and targeted hedging effects.  
& ``Between Condition'' and  
``Hedging Mechanism'';  
\cref{tab:s4_pairwise} and \cref{tab:s4_mechanism}. \\[3pt]  
  
Exact dialogue-level sign-flip test  
& Check the specified oracle-hedging mechanism contrast across  
source dialogues.  
& ``Hedging Mechanism'';  
\cref{tab:s4_mechanism};  
\cref{sec:stats}. \\  
  
\midrule  
\multicolumn{3}{@{}l}{\textit{Additional inferential procedures (not standalone tests)}} \\  
\midrule  
  
Percentile bootstrap confidence intervals  
& Quantify uncertainty in calibration metrics and human-study  
effects/AUC; typically $10{,}000$ resamples.  
& \cref{sec:study1},  
and \cref{sec:stats};  
\cref{tab:s1_calibration_full}. \\[3pt]  
  
Holm multiple-testing correction  
& Control family-wise error separately for primary and exploratory  
human-study comparison families.  
& \cref{sec:study_results};  
Tables~\ref{tab:s4_pairwise} and~\ref{tab:s4_mechanism};  
\cref{sec:stats}. \\  
  
\bottomrule  
\end{tabular}  
\end{table}  

\end{document}